\documentclass{article}

\usepackage[preprint]{preprint}
\pdfoutput=1

\usepackage[utf8]{inputenc} 
\usepackage[T1]{fontenc}    
\usepackage{hyperref}       
\usepackage{url}            
\usepackage{booktabs}       
\usepackage{amsfonts}       
\usepackage{nicefrac}       
\usepackage{microtype}      

\usepackage[ruled,vlined,linesnumbered]{algorithm2e}
\usepackage{xspace}
\usepackage{booktabs}
\usepackage{colortbl}
\usepackage{mathtools}
\usepackage{amsmath}
\usepackage{amssymb}
\usepackage{amsthm}
\usepackage{amsfonts}
\usepackage{cleveref}
\Crefformat{section}{Section~#2#1#3}
\Crefformat{proposition}{Proposition~#2#1#3}
\Crefformat{equation}{Eq.\,#2#1#3} 

\usepackage{bm}
\usepackage{xspace}
\usepackage{enumitem}
\usepackage[dvipsnames]{xcolor}
\usepackage{subcaption}
\usepackage{caption}
\usepackage[subpreambles=true]{standalone}
\usepackage{wrapfig}
\usepackage{stackengine}
\usepackage{multirow}
\usepackage{array}
\usepackage{wrapfig}
\usepackage[final]{pdfpages}

\newcolumntype{P}[1]{>{\centering\arraybackslash}p{#1}}

\usepackage[colorinlistoftodos, textwidth=30mm, shadow, textsize=small]{todonotes}
\definecolor{babyblue}{rgb}{0.54, 0.81, 0.94}
\definecolor{citrine}{rgb}{0.89, 0.82, 0.04}
\definecolor{misocolor}{rgb}{0.16,0.27,0.86}
\definecolor{jbcolor}{rgb}{0.9,0.4,0.2}
\definecolor{bernacolor}{rgb}{0.9608,0.4863,0.00}
\definecolor{carlcolor}{rgb}{0.0,0.9863,0.30}
\definecolor{grey}{rgb}{0.3, 0.3, 0.3}

\definecolor{graphicbackground}{rgb}{0.96,0.96,0.8}
\definecolor{rouge1}{RGB}{226,0,38}  
\definecolor{orange1}{RGB}{243,154,38}  
\definecolor{jaune}{RGB}{254,205,27}  
\definecolor{blanc}{RGB}{255,255,255} 
\definecolor{rouge2}{RGB}{230,68,57}  
\definecolor{orange2}{RGB}{236,117,40}  
\definecolor{taupe}{RGB}{134,113,127} 
\definecolor{gris}{RGB}{91,94,111} 
\definecolor{bleu1}{RGB}{38,109,131} 
\definecolor{bleu2}{RGB}{28,50,114} 
\definecolor{vert1}{RGB}{133,146,66} 
\definecolor{vert3}{RGB}{20,200,66} 
\definecolor{vert2}{RGB}{157,193,7} 
\definecolor{darkyellow}{RGB}{233,165,0}  
\definecolor{lightgray}{rgb}{0.9,0.9,0.9}
\definecolor{darkgray}{rgb}{0.6,0.6,0.6}
\definecolor{babyblue}{rgb}{0.54, 0.81, 0.94}
\definecolor{citrine}{rgb}{0.89, 0.82, 0.04}
\definecolor{misogreen}{rgb}{0.25,0.6,0.0}
\definecolor{PalePurp}{rgb}{0.66,0.57,0.66}
\definecolor{todocolor}{rgb}{0.66,0.99,0.99}
\definecolor{pearOne}{HTML}{2C3E50}
\definecolor{pearTwo}{HTML}{A9CF54}
\definecolor{pearTwoT}{HTML}{C2895B}
\definecolor{pearThree}{HTML}{E74C3C}
\colorlet{titleTh}{pearOne}
\colorlet{bull}{pearTwo}
\definecolor{pearcomp}{HTML}{B97E29}
\definecolor{pearFour}{HTML}{588F27}
\definecolor{pearFith}{HTML}{ECF0F1}
\definecolor{pearDark}{HTML}{2980B9}
\definecolor{pearDarker}{HTML}{1D2DEC}

\hypersetup{
	colorlinks,
	citecolor=pearDark,
	linkcolor=pearThree,
breaklinks=true,
	urlcolor=pearDarker}

\DeclareMathOperator*{\argmin}{arg\,min}

\DeclareMathOperator*{\var}{Var}

\newcommand{\simclr}{\texttt{SimCLR}\xspace}
\newcommand{\meanteacher}{\texttt{Mean Teacher}\xspace}
\newcommand{\moco}{\texttt{MoCo}\xspace}

\newcommand{\BN}{\normalfont\textsc{B\hskip -0.1em N}\xspace}
\newcommand{\LN}{\normalfont\textsc{L\hskip -0.1em N}\xspace}
\newcommand{\Id}{\normalfont\textsc{I\hskip -0.1em d}\xspace}
\newcommand{\BNt}{\normalfont\textsc{BatchNorm}\xspace}
\newcommand{\LNt}{\normalfont\textsc{LayerNorm}\xspace}

\newcommand{\LARS}{\normalfont\texttt{LARS}\xspace}
\newcommand{\MT}{\normalfont\texttt{MT}\xspace}
\newcommand{\SGD}{\normalfont\texttt{SGD}\xspace}
\newcommand{\LBFGS}{\normalfont\texttt{LBFGS}\xspace}

\newcommand{\CMC}{\normalfont\texttt{CMC}\xspace}
\newcommand{\CPCC}{\normalfont\texttt{CPC v2}\xspace}
\newcommand{\PIRL}{\normalfont\texttt{PIRL}\xspace}

\newcommand{\bOne}{{\bf 1}}

\newcommand{\EE}[1]{\mathbb{E}\left[#1\right]}

\newcommand{\pa}[1]{\left(#1\right)}

\let\originalleft\left
\let\originalright\right
\renewcommand{\left}{\mathopen{}\mathclose\bgroup\originalleft}
\renewcommand{\right}{\aftergroup\egroup\originalright}

\newcommand{\normtwo}[1]{\left\|#1\right\|_2}

\newcommand{\CommaBin}{\mathbin{\raisebox{0.5ex}{,}}}
\newcommand*{\eqdef}{\triangleq}
\newcommand{\transpose}{^\mathsf{\scriptscriptstyle T}}

\renewcommand{\epsilon}{\varepsilon}
\renewcommand{\tilde}{\widetilde}
\renewcommand{\bar}{\overline}

\newcommand{\nothere}[1]{}

\usepackage{xspace}

\newcommand{\algo}{\texttt{BYOL}\xspace}
\newcommand{\augview}{augmented view}
\newcommand{\aaugview}{an augmented view}
\newcommand{\imageaugmentation}{image augmentation}
\newcommand{\Imageaugmentation}{Image augmentation}

\newcommand{\dtheta}{{\delta\hspace{-.1em}\theta}}
\newcommand{\supervisedbaseline}{Supervised-IN}
\newcommand{\longalgo}{\textbf{B}ootstrap \textbf{Y}our \textbf{O}wn \textbf{L}atent}
\newcommand{\netparams}{{\theta}}
\newcommand{\targetparams}{\xi}
\renewcommand{\eqdef}{\mathrel{\ensurestackMath{\stackon[1pt]{=}{\scriptscriptstyle\Delta}}}}

\title{Bootstrap Your Own Latent\\A New Approach to Self-Supervised Learning}

\author{%
Jean-Bastien Grill\thanks{Equal contribution; the order of first authors was randomly selected.} $^{ , 1}$ \; 
Florian Strub\footnotemark[1] $^{ , 1}$ \;
Florent Altché\footnotemark[1] $^{ , 1}$ \;
Corentin Tallec\footnotemark[1] $^{ , 1}$ \;
Pierre H.\,Richemond\footnotemark[1] $^{ ,1, 2}$ \AND
Elena Buchatskaya$^{1}$  \;
Carl Doersch$^{1}$  \;
Bernardo Avila Pires$^{1}$  \;
Zhaohan Daniel Guo$^{1}$ \AND
Mohammad Gheshlaghi Azar$^{1}$ \;
Bilal Piot$^{1}$ \;
Koray Kavukcuoglu$^{1}$ \;
Rémi Munos$^{1}$ \;
Michal Valko$^{1}$ \vspace{0.7em}\\
$^{1}$DeepMind \hspace{1cm} $^{2}$Imperial College\vspace{.7em}\\
\texttt{[jbgrill,fstrub,altche,corentint,richemond]@google.com}
}

\begin{document}

\maketitle

\begin{abstract}
We introduce \longalgo~(\algo), a new approach to self-supervised image representation learning. \algo relies on two neural networks, referred to as \textit{online} and \textit{target} networks, that interact and learn from each other. 
From an \augview~of an image, we train the online network to predict the target network representation of the same image under a different \augview. 
At the same time, we update the target network with a slow-moving average of the online network. 
    While state-of-the art methods rely on negative pairs, \algo achieves a new state of the art \textit{without them}. \algo reaches $74.3\%$ top-1 classification accuracy on ImageNet using a linear evaluation with a ResNet-50 architecture and $79.6\%$ with a larger ResNet. 
We show that \algo performs on par or better than the current state of the art on both transfer and semi-supervised benchmarks. Our implementation and pretrained models are given on GitHub.\setcounter{footnote}{2}\footnote{\url{https://github.com/deepmind/deepmind-research/tree/master/byol}}
\end{abstract}

\section{Introduction}
\label{sec:introduction}
\vspace{-1.3em}
\phantom{\cite{fukushima1980neocognitron, wiskott2002slow, hinton2006fast,oquab2014learning, simonyan2014very, girshick2014rich, long2015fully}}

\begin{wrapfigure}{r}{0.50\textwidth}
  \vspace{-7.7mm}
   \centering
   \begin{minipage}{0.5\textwidth}
    \centering
     \includegraphics[width=0.95\textwidth,bb=0 0 1140 1018]{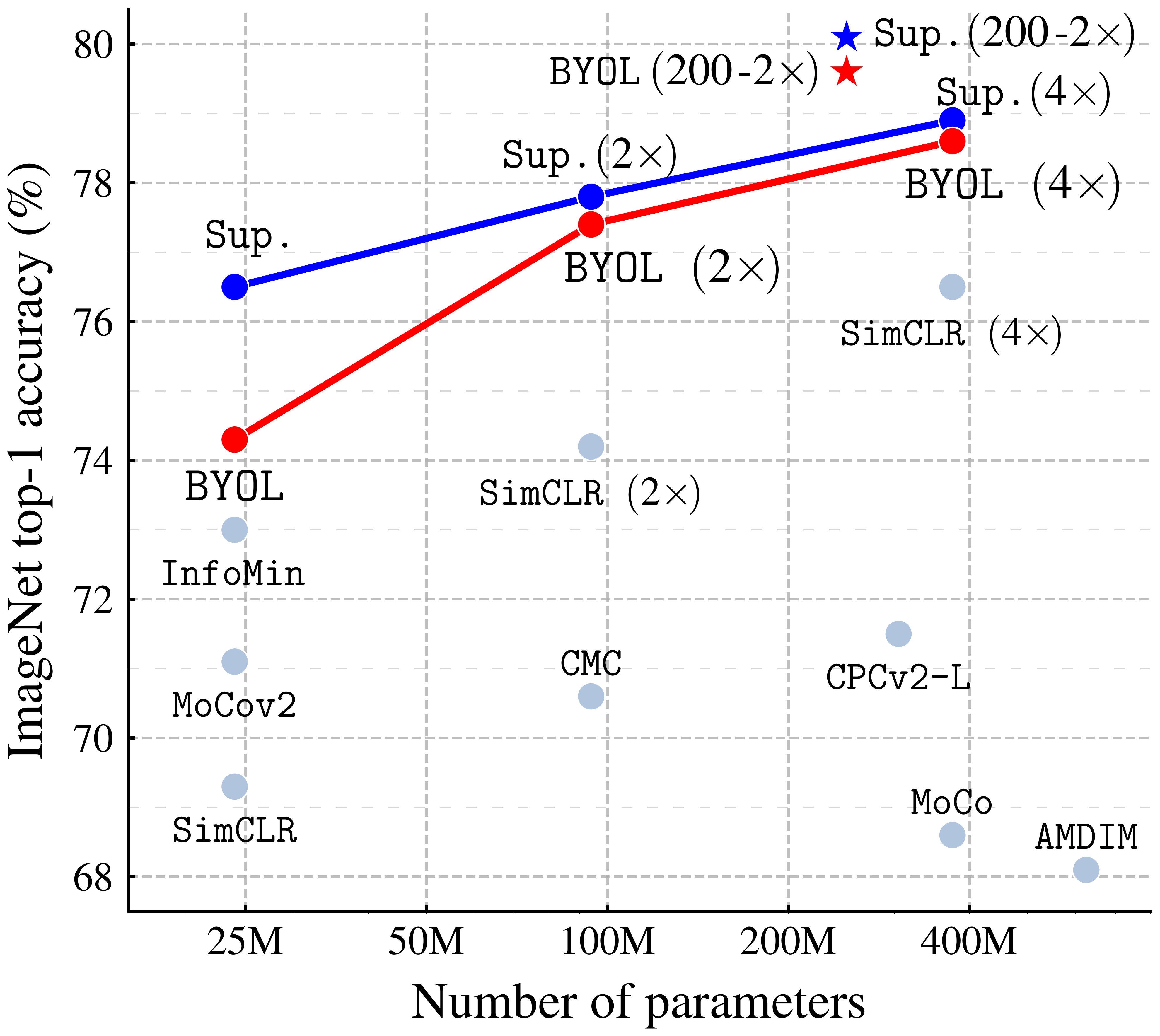}
        \caption{Performance of \algo on ImageNet (linear evaluation) using ResNet-50 and our best architecture ResNet-200 ($2\times$), compared to other unsupervised and supervised (\texttt{Sup\!.\!}) baselines~\cite{SimCLR}.}
        \label{fig:holygrail}
 \end{minipage}
 \vspace{-8mm}
\end{wrapfigure}

Learning good image representations is a key challenge in computer vision~\cite{fukushima1980neocognitron, wiskott2002slow, hinton2006fast} as it allows for efficient training on downstream tasks~\cite{oquab2014learning, simonyan2014very, girshick2014rich, long2015fully}.
Many different training approaches have been proposed to learn such representations, usually relying on visual pretext tasks. 
Among them, state-of-the-art contrastive methods~\cite{SimCLR, MOCO, CPC, CMC, Tian2020WhatMF} are trained by reducing the distance between representations of different \augview s of the same image (`positive pairs'), and increasing the distance between representations of \augview s from different images (`negative pairs'). These methods need careful treatment of negative pairs~\cite{pmlr-v97-saunshi19a} by either relying on large batch sizes~\cite{SimCLR, Tian2020WhatMF}, memory banks~\cite{MOCO} or customized mining strategies~\cite{Manmatha2017SamplingMI, Harwood2017SmartMF} to retrieve the negative pairs. In addition, their performance critically depends on the choice of \imageaugmentation s~\cite{SimCLR,Tian2020WhatMF}. 

In this paper, we introduce \longalgo~(\algo), a new algorithm for self-supervised learning of image representations. \algo achieves higher performance than state-of-the-art contrastive methods without using negative pairs. It iteratively bootstraps\footnote{Throughout this paper, the term \emph{bootstrap} is used in its idiomatic sense rather than the statistical sense.} the outputs of a network to serve as targets for an enhanced representation.
Moreover, \algo is more robust to the choice of \imageaugmentation s than contrastive methods; we suspect that not relying on negative pairs is one of the leading reasons for its improved robustness. While previous methods based on bootstrapping have used pseudo-labels~\cite{lee2013PseudoLabelT}, cluster indices~\cite{caron2018DeepCF} or a handful of labels~\cite{bachman2014learning, laine2016temporal, tarvainen2017mean}, 
we propose to \textit{directly bootstrap the representations}. In particular, \algo uses two neural networks, referred to as online and target networks, that interact and learn from each other. Starting from \aaugview~of an image, \algo trains its online network to predict the target network's representation of another \augview~of the same image. 
While this objective admits collapsed solutions, e.g., outputting the same vector for all images, we empirically show that \algo does not converge to such solutions. 
We hypothesize (see~\Cref{sec:intuition_byol}) that the combination of
(i) the addition of a predictor to the online network 
and (ii) the use of a slow-moving average of the online parameters as the target network encourages encoding more and more information within the online projection and avoids collapsed solutions.

We evaluate the representation learned by \algo on ImageNet~\cite{ILSVRC15} and other vision benchmarks using ResNet architectures~\cite{ResNet}. Under the linear evaluation protocol on ImageNet, consisting in training a linear classifier on top of the frozen representation, \algo reaches $74.3\%$ top-1 accuracy with a standard ResNet-$50$ and $79.6\%$ top-1 accuracy with a larger ResNet (\Cref{fig:holygrail}). 
In the semi-supervised and transfer settings on ImageNet, we obtain results on par or superior to the current state of the art. Our contributions are: (\textit{i}) We introduce \algo, a self-supervised representation learning method (\Cref{sec:method}) which achieves state-of-the-art results under the linear evaluation protocol on ImageNet without using negative pairs. (\textit{ii}) We show that our learned representation outperforms the state of the art on semi-supervised and transfer benchmarks (\Cref{sec:experiments}). (\textit{iii}) We show that \algo is more resilient to changes in the batch size and in the set of \imageaugmentation s compared to its contrastive counterparts (\Cref{sec:ablation}).
In particular, \algo suffers a much smaller performance drop than \simclr, a strong contrastive baseline, when only using random crops as \imageaugmentation s.

\section{Related work}
\label{sec:related_work}
Most unsupervised methods for representation learning can be categorized as either generative or discriminative~\cite{doersch2015unsupervised, SimCLR}. 
Generative approaches to representation learning build a distribution over data and latent embedding and use the learned embeddings as image representations. Many of these approaches rely either on auto-encoding of images~\cite{vincent2008extracting, KingmaVAE, RezendeVAE} or on adversarial learning~\cite{GAN}, jointly modelling data and representation~\cite{donahue2016adversarial, DumoulinALI, BigBiGAN, BrockBigGAN}. Generative methods typically operate directly in pixel space. This however is computationally expensive, and the high level of detail required for image generation may not be necessary for representation learning.

Among discriminative methods, contrastive methods~\cite{MOCO, CPC, CPCv2, DIM, AMDIM, CMC, PIRL, li2020prototypical}
currently achieve state-of-the-art performance in self-supervised learning~\cite{MOCOv2, SimCLR,chen2020big,Tian2020WhatMF}. Contrastive approaches avoid a costly generation step in pixel space by bringing representation of different views of the same image closer (`positive pairs'), and spreading representations of views from different images (`negative pairs') apart~\cite{wu2018unsupervised, doersch2017multi}. Contrastive methods often require comparing each example with many other examples to work well~\cite{MOCO, SimCLR} prompting the question of whether using negative pairs is necessary. 

\texttt{DeepCluster}~\cite{caron2018DeepCF} partially answers this question. It uses bootstrapping on previous versions of its representation to produce targets for the next representation; it clusters data points using the prior representation, and uses the cluster index of each sample as a classification target for the new representation. While avoiding the use of negative pairs, this requires a costly clustering phase and specific precautions to avoid collapsing to trivial solutions. 

Some self-supervised methods are not contrastive but rely on using auxiliary handcrafted prediction tasks to learn their representation. In particular, relative patch prediction \cite{doersch2015unsupervised,doersch2017multi}, colorizing gray-scale images~\cite{Zhang2016ColorfulIC, larsson2016learning}, image inpainting~\cite{pathak2016context}, image jigsaw puzzle~\cite{noroozi2016unsupervised}, image super-resolution~\cite{ledig2017photo}, and geometric transformations~\cite{dosovitskiy2014discriminative, gidaris2018unsupervised} have been shown to be useful. Yet, even with suitable architectures~\cite{kolesnikov2019revisiting}, these methods are being outperformed by contrastive methods~\cite{MOCOv2, SimCLR, Tian2020WhatMF}.

Our approach has some similarities with \textit{Predictions of Bootstrapped Latents} (\texttt{PBL}, \cite{guo2020bootstrap}), a self-supervised representation learning technique for reinforcement learning (RL). \texttt{PBL} jointly trains the agent's history representation and an encoding of future observations. The observation encoding is used as a target to train the agent's representation, and the agent's representation as a target to train the observation encoding. Unlike \texttt{PBL}, \algo uses a slow-moving average of its representation to provide its targets, and does not require a second network.

The idea of using a slow-moving average target network to produce stable targets for the online network was inspired by  deep RL \cite{DQNNature, mnih2016asynchronous, hessel2018rainbow, van2018deep}. Target networks stabilize the bootstrapping updates provided by the Bellman equation, making them appealing to stabilize the bootstrap mechanism in \algo. While most RL methods use fixed target networks, \algo uses a weighted moving average of previous networks (as in \cite{DDPG}) in order to provide smoother changes in the target representation. 

In the semi-supervised setting~\cite{chapelle2009semi,zhu2009introduction}, an unsupervised loss is combined with a classification loss over a handful of labels to ground the training~\cite{laine2016temporal,tarvainen2017mean,kingma2014semi,rasmus2015semi,berthelot2019mixmatch,miyato2018virtual,Berthelot2020ReMixMatchSL,sohn2020fixmatch}. Among these methods, \textit{mean teacher} (\MT) \cite{tarvainen2017mean} also uses a slow-moving average network, called \textit{teacher}, to produce targets for an online network, called \textit{student}. 
An $\ell_2$ consistency loss between the softmax predictions of the teacher and the student is added to the classification loss.
While \cite{tarvainen2017mean} demonstrates the effectiveness of \MT in the semi-supervised learning case, in~\Cref{sec:mean_teacher} we show that a similar approach collapses when removing the classification loss. In contrast, \algo introduces an additional predictor on top of the online network, which prevents collapse. 

Finally, in self-supervised learning, \moco~\cite{MOCO} uses a slow-moving average network (\emph{momentum encoder}) to maintain consistent representations of negative pairs drawn from a memory bank. Instead, \algo uses a moving average network to produce prediction targets as a means of stabilizing the bootstrap step. We show in \Cref{subsec:simclr_to_pbl} that this mere stabilizing effect can also improve existing contrastive methods.

\section{Method}
\label{sec:method}
We start by motivating our method before explaining its details in \Cref{subsec:desc_algo}. Many successful self-supervised learning approaches build upon the cross-view prediction framework introduced in~\cite{becker1992self}. Typically, these approaches learn representations  by predicting different views (e.g., different random crops) 
of the same image from one another. Many such  approaches cast the prediction problem directly in representation space: the representation of an  \augview~of an image should be predictive of the representation of another \augview~of the same image. However, predicting directly in representation space can lead to collapsed representations: for instance, a representation that is constant across views is always fully predictive of itself. Contrastive methods circumvent this problem by reformulating the prediction problem into one of discrimination: from the representation of an \augview, they learn to discriminate between the representation of another \augview~of the same image, and the representations of \augview s of different images. In the vast majority of cases, this prevents the training from finding collapsed representations. Yet, this discriminative approach typically requires comparing each representation of an \augview~with many negative examples, to find ones sufficiently close to make the discrimination task challenging. In this work, we thus tasked ourselves to find out whether these negative examples are indispensable to prevent collapsing while preserving high performance.

To prevent collapse, a straightforward solution is to use a fixed randomly initialized network to produce the targets for our predictions. While avoiding collapse, it empirically does not result in very good representations. Nonetheless, it is interesting to note that the representation obtained using this procedure can already be much better than the initial fixed representation. In our ablation study (\Cref{sec:ablation}), we apply this procedure by predicting a fixed randomly initialized network and achieve $18.8\%$ top-1 accuracy (\Cref{tab:ablation_bootstrap}) on the linear evaluation protocol on ImageNet, whereas the randomly initialized network only achieves $1.4\%$ by itself. This experimental finding is the core motivation for \algo: from a given representation, referred to as \textit{target}, we can train a new, potentially enhanced representation, referred to as \textit{online}, by predicting the target representation. From there, we can expect to build a sequence of representations of increasing quality by iterating this procedure, using subsequent online networks as new target networks for further training. In practice, \algo generalizes this bootstrapping procedure by iteratively refining its representation, but using a slowly moving exponential average of the online network as the target network instead of fixed checkpoints.

\subsection{Description of \algo}
\label{subsec:desc_algo}
\begin{figure}[!ht]
\vskip -35em
    \centering
        \includegraphics[bb=0 0  1233 461]{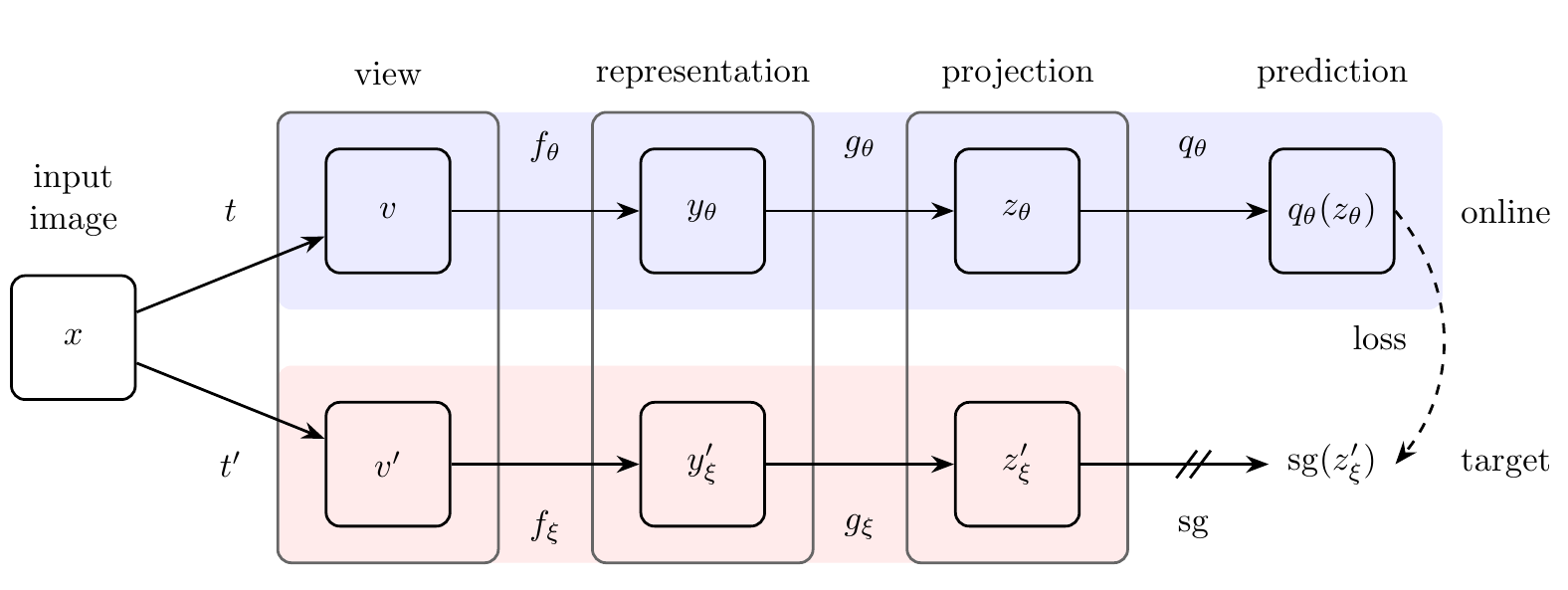}
    \caption{\algo's architecture. \algo minimizes a similarity loss between $q_\netparams(z_\netparams)$ and $\mathrm{sg}(z'_\targetparams)$, where $\netparams$ are the trained weights, $\targetparams$ are an exponential moving average of $\netparams$ and $\mathrm{sg}$ means stop-gradient. At the end of training, everything but $f_\netparams$ is discarded, and $y_\netparams$ is used as the image representation. }
    \label{fig:algo}
\end{figure}
\algo's goal is to learn a representation $y_\netparams$ which can then be used for downstream tasks. As described previously, \algo uses two neural networks to learn: the \emph{online} and \emph{target} networks. The online network is defined by a set of weights $\netparams$ and is comprised of three stages: an \emph{encoder} $f_\netparams$, a \emph{projector}~$g_\netparams$ and a \emph{predictor}~$q_\netparams$, as shown in \Cref{fig:algo} and \Cref{fig:BYOL_sketch}. The target network has the same architecture as the online network, but uses a different set of weights $\targetparams$.
The target network provides the regression targets to train the online network, and its parameters $\targetparams$ are an exponential moving average of the online parameters $\netparams$~\cite{DDPG}. More precisely, given a target decay rate $\tau \in [0, 1]$, after each training step we perform the following update,
\begin{equation}
\targetparams \leftarrow \tau\targetparams + (1 - \tau) \netparams.  \label{eq:tau}  
\end{equation}
Given a set of images $\mathcal{D}$, an image $x \sim \mathcal{D}$ sampled uniformly from $\mathcal{D}$, and two distributions of \imageaugmentation s $\mathcal{T}$ and $\mathcal{T}'$, \algo produces two \augview s $v \eqdef t(x)$ and $v'\eqdef t'(x)$ from $x$ by applying respectively \imageaugmentation s $t\sim \mathcal{T}$ and $t'\sim \mathcal{T}'$. From the first \augview~$v$, the online network outputs a \emph{representation} $y_\netparams \eqdef f_\netparams(v)$ and a projection $z_\netparams \eqdef g_\netparams(y)$. The target network outputs $y'_\targetparams \eqdef f_\targetparams(v')$ and the \emph{target projection} $z'_\targetparams \eqdef g_\targetparams (y')$ from the second \augview~$v'$. We then output a \emph{prediction} $q_\netparams(z_\netparams)$ of $z'_\targetparams$ and $\ell_2$-normalize both $q_\netparams(z_\netparams)$ and $z'_\targetparams$ to $\bar{q_\netparams}(z_\netparams) \eqdef q_\netparams(z_\netparams)/\normtwo{q_\netparams(z_\netparams)}$ and $\bar{z}'_\targetparams \eqdef z'_\targetparams / \|z'_\targetparams\|_2$.  Note that this predictor is only applied to the online branch, making the architecture asymmetric between the online and target pipeline. Finally we define the following mean squared error between the normalized predictions and target projections,\footnote{While we could directly predict the representation~$y$ and not a projection $z$, previous work~\cite{SimCLR} have empirically shown that using this projection improves performance.}
\begin{equation}
    \mathcal{L}_{\netparams, \targetparams} \eqdef \normtwo{\bar{q_{\theta}}(z_\netparams) - \bar{z}'_\targetparams}^2 = 2 - 2\cdot\frac{\langle q_\netparams(z_\netparams),  z'_\targetparams \rangle }{\big\|q_\netparams(z_\netparams)\big\|_2\cdot
    \big\|z'_\targetparams\big\|_2
    }
    \cdot
    \label{eq:cosine-loss}
\end{equation}

We symmetrize the loss $\mathcal{L}_{\netparams, \targetparams}$ in \Cref{eq:cosine-loss} by separately feeding $v'$ to the online network and $v$ to the target network to compute $\tilde{\mathcal{L}}_{\netparams, \targetparams}$. At each training step, we perform a stochastic optimization step to minimize $\mathcal{L}^\algo_{\netparams, \targetparams} = \mathcal{L}_{\netparams, \targetparams} + \tilde{\mathcal{L}}_{\netparams, \targetparams}$ with respect to $\netparams$ only, but \textit{not} $\targetparams$, as depicted by the stop-gradient in \Cref{fig:algo}. \algo's dynamics are summarized as
\begin{align}
&\netparams \leftarrow \mathrm{optimizer} \left(\netparams, \nabla_\netparams \mathcal{L}^\algo_{\netparams, \targetparams}, \eta \right),
\\
&\targetparams \leftarrow \tau \targetparams + (1 - \tau) \netparams,  \tag{\ref{eq:tau}}
\end{align}
where $\mathrm{optimizer}$ is an optimizer and $\eta$ is a learning rate.

At the end of training, we only keep the encoder $f_\netparams$; as in~\cite{MOCO}. When comparing to other methods, we consider the number of inference-time weights only in the final representation $f_\netparams$.
The full training procedure is summarized in \Cref{app:algo}, and \texttt{python} pseudo-code based on the libraries \texttt{JAX}~\cite{jax2018github} and \texttt{Haiku}~\cite{haiku2020github} is provided in
in Appendix~\hyperlink{page.33}{J}.

\subsection{Intuitions on \algo's behavior}
\label{sec:intuition_byol}
As \algo does not use an explicit term to prevent collapse (such as negative examples~\cite{CPC}) while minimizing $\mathcal L_{\netparams,\targetparams}^\algo$ with respect to $\netparams$, it may seem that \algo should converge to a minimum of this loss with respect to $(\netparams, \targetparams)$ (\emph{e.g.}, a collapsed constant representation). However \algo's target parameters $\targetparams$ updates are \textbf{not} in the direction of $\nabla_\targetparams \mathcal L_{\netparams,\targetparams}^\algo$. More generally, we hypothesize that there is no loss $L_{\netparams, \targetparams}$ such that \algo's dynamics is a gradient descent on $L$ jointly over $\netparams, \targetparams$. This is similar to GANs~\cite{goodfellow2014generative}, where there is no loss that is jointly minimized w.r.t.\,both the discriminator and generator parameters. 
There is therefore no \textit{a priori} reason why \algo's parameters would converge to a minimum of $\mathcal{L}^\algo_{\netparams, \targetparams}$. 

While \algo's dynamics still admit undesirable equilibria, we did not observe convergence to such equilibria in our experiments. In addition, when assuming \algo's predictor to be optimal\footnote{For simplicity we also consider \algo without normalization (which performs reasonably close to \algo, see \Cref{app:abla_norm}) nor symmetrization} i.e., $q_\netparams = q^\star$ with 

\begin{equation}
\label{eq:optimal_predictor}
    q^\star \eqdef \argmin_q \EE{\normtwo{q(z_\netparams) - z'_\targetparams}^2}, \quad \text{where} \quad q^\star(z_\netparams) = \EE{z'_\targetparams | z_\netparams},
\end{equation}

we hypothesize that the undesirable equilibria are unstable. Indeed, in this optimal predictor case, \algo's updates on~$\netparams$ follow in expectation the gradient of the expected conditional variance (see \Cref{app:thm-envelope} for details),
\begin{equation}
\label{eq:byol_minimize_cond_var}
    \nabla_\netparams\EE{\normtwo{q^\star(z_\netparams) - z'_\targetparams}^2} = \nabla_\netparams\EE{\normtwo{\EE{z'_\targetparams | z_\netparams} -z'_\targetparams }^2} = \nabla_\netparams \EE{ \sum_i \var(z'_{\targetparams, i} | z_\netparams)},
\end{equation}
where $z'_{\targetparams, i}$ is the $i$-th feature of $z'_{\targetparams}$. 

Note that for any random variables $X,$ $Y,$ and $Z$, $\var(X|Y,Z) \leq \var(X|Y)$. Let $X$ be the target projection, $Y$ the current online projection, and $Z$ an additional variability on top of the online projection induced by stochasticities in the training dynamics: purely discarding information from the online projection cannot decrease the conditional variance. 

In particular, \algo avoids constant features in $z_\netparams$ 
as, for any constant $c$ and random variables $z_\netparams$ and $z'_\targetparams$, $\var(z'_\targetparams | z_\netparams) \leq \var(z'_\targetparams | c)$;  hence our hypothesis on these collapsed constant equilibria being unstable. Interestingly, if we were to minimize $\mathbb{E}[\sum_i \var(z'_{\targetparams, i} | z_\netparams)]$ with respect to $\targetparams$, we would get a collapsed $z'_\targetparams$ as the variance is minimized for a constant $z'_\targetparams$. 
Instead, \algo makes $\targetparams$ closer to $\netparams$, incorporating sources of variability captured by the online projection into the target projection. 

Furthemore, notice that performing a hard-copy of the online parameters $\netparams$ into the target parameters $\targetparams$ would be enough to propagate new sources of variability. However, sudden changes in the target network might break the assumption of an optimal predictor, in which case \algo's loss is not guaranteed to be close to the conditional variance. We hypothesize that the main role of \algo's moving-averaged target network is to ensure the near-optimality of the predictor over training; \Cref{sec:ablation} and \Cref{app:importance_no_pred} provide some empirical support of this interpretation.

\subsection{Implementation details}
\label{sec:impl_detail}
\paragraph{\Imageaugmentation s} \algo uses the same set of \imageaugmentation s as in \simclr~\cite{SimCLR}. First, a random patch of the image is selected and resized to $224 \times 224$ with a random horizontal flip, followed by a color distortion, consisting of a random sequence of brightness, contrast, saturation, hue adjustments, and an optional grayscale conversion. Finally Gaussian blur and solarization are applied to the patches. Additional details on the \imageaugmentation s are in \Cref{app:transf}.
\paragraph{Architecture} We use a convolutional residual network \cite{ResNet} with 50 layers and post-activation (ResNet-$50(1 \times)$ v1) as our base parametric encoders $f_\netparams$ and $f_\targetparams$. We also use deeper ($50$, $101$, $152$ and $200$ layers) and wider (from $1 \times$ to $4 \times$) ResNets, as in \cite{zagoruyko2016wide,kolesnikov2019revisiting,SimCLR}. Specifically, the representation $y$ corresponds to the output of the final average pooling layer, which has a feature dimension of $2048$ (for a width multiplier of $1\times$). As in \simclr~\cite{SimCLR}, the representation~$y$ is projected to a smaller space by a \emph{multi-layer perceptron} (MLP)~$g_\netparams$, and similarly for the target projection $g_\targetparams$. This MLP consists in a linear layer with output size $4096$ followed by batch normalization~\cite{BatchNorm}, rectified linear units (ReLU)~\cite{nair2010rectified}, and a final linear layer with output dimension $256$. Contrary to \simclr, the output of this MLP is not batch normalized. The predictor $q_\netparams$ uses the same architecture as $g_\netparams$.

\paragraph{Optimization}
We use the \LARS optimizer~\cite{LARS} with a cosine decay learning rate schedule~\cite{SGDR}, without restarts, over $1000$ epochs, with a warm-up period of $10$ epochs. We set the base learning rate to $0.2,$ scaled linearly~\cite{Goyal2017} with the batch size ($\text{LearningRate} = 0.2\times \text{BatchSize} / 256$). In addition, we use a global weight decay parameter of $1.5\cdot10^{-6}$ while excluding the biases and batch normalization parameters from both \LARS adaptation and weight decay. For the target network, the exponential moving average parameter $\tau$ starts from $\tau_\text{base} = 0.996$ and is increased to one during training. Specifically, we set $\tau \triangleq 1 - (1-\tau_\text{base})\cdot \pa{\cos\pa{\pi k / K} + 1} / 2$ with $k$ the current training step and $K$ the maximum number of training steps. We use a batch size of $4096$ split over $512$ Cloud TPU v$3$ cores. With this setup, training takes approximately $8$ hours for a ResNet-$50(\times 1)$. All hyperparameters are summarized in Appendix~\hyperlink{page.33}{J}; 
an additional set of hyperparameters for a smaller batch size of $512$ is provided in \Cref{app:smaller-batch}.

\section{Experimental evaluation}
\label{sec:experiments}
\vspace{-0.4em}
We assess the performance of \algo 's representation 
after self-supervised pretraining on the training set of the ImageNet ILSVRC-2012 dataset~\cite{ILSVRC15}. We first evaluate it on ImageNet (IN) in both linear evaluation and semi-supervised setups. We then measure its transfer capabilities on other datasets and tasks, including classification, segmentation, object detection and depth estimation. For comparison, we also report scores for a representation trained using labels from the \texttt{train\xspace} ImageNet subset, referred to as \supervisedbaseline. In \Cref{app:train-places}, we assess the generality of \algo by pretraining a representation on the Places365-Standard dataset~\cite{zhou2017places} before reproducing this evaluation protocol.

\vfill
\paragraph{Linear evaluation on ImageNet}
We first evaluate \algo's representation by training a linear classifier on top of the frozen representation, following the procedure described in~\cite{kolesnikov2019revisiting,kornblith2019better,Zhang2016ColorfulIC,CPC,SimCLR}, and \cref{app:linear_eval_details}; we report top-$1$ and top-$5$ accuracies in \% on the \texttt{test\xspace} set in \Cref{table:linear_classif}. With a standard ResNet-$50$ ($\times 1$) \algo obtains $74.3\%$ top-$1$ accuracy ($91.6\%$ top-$5$ accuracy), which is a $1.3\%$ (resp.\,$0.5\%$) improvement over the previous self-supervised state of the art~\cite{Tian2020WhatMF}. This tightens the gap with respect to the supervised baseline of~\cite{SimCLR}, $76.5\%$, but is still significantly below the stronger supervised baseline of~\cite{gong2020maxup}, $78.9\%$. With deeper and wider architectures, \algo consistently outperforms the previous state of the art (\Cref{app:larger_architectures}), and obtains a best performance of $79.6\%$ top-$1$ accuracy, ranking higher than previous self-supervised approaches. On a ResNet-$50$ ($4\times$) \algo achieves $78.6\%$, similar to the $78.9\%$ of the best supervised baseline in \cite{SimCLR} for the same architecture. 

\vfill

\begin{table}[h]
\begin{subtable}{.38\linewidth}
    \small
    \centering
    \begin{tabular}[t]{l l r r r}
    \toprule
         Method &  Top-$1$ & Top-$5$ \\
         \midrule
         Local Agg.\,& $60.2$ & -\\
         \PIRL \cite{PIRL} & $63.6$ & -\\
         \CPCC \cite{CPCv2} & $63.8$ & $85.3$\\
         \CMC \cite{CMC} & $66.2$ & $87.0$\\
         \simclr \cite{SimCLR} & $69.3$ & $89.0$\\
         \moco v2 \cite{MOCOv2} & $71.1$ & -\\
         InfoMin Aug.\,\cite{Tian2020WhatMF} & $73.0$ & $91.1$\\
         \algo (ours) & $\cellcolor{pearDark!20}\bf{74.3}$ & $\cellcolor{pearDark!20}\bf{91.6}$\\
         \bottomrule
\end{tabular}
     \caption{\label{table:linear_classif_1}ResNet-50 encoder.}
\end{subtable}
\hfill
\begin{subtable}{.62\linewidth}
     \small
    \centering
    \begin{tabular}[t]{l l r r r}
    \toprule
    Method & Architecture & Param. &  Top-$1$ & Top-$5$ \\
    \midrule 
         \simclr \cite{SimCLR}      & ResNet-$50$ ($2\times$)  & $94$M  & $74.2$ & $92.0$\\
         \CMC \cite{CMC}            & ResNet-$50$ ($2\times$)  & $94$M  & $70.6$ & $89.7$\\
         \algo (ours)               & ResNet-$50$ ($2\times$)  & $94$M  & \cellcolor{pearDark!20} $77.4$ & \cellcolor{pearDark!20} $93.6$\\
         \CPCC \cite{CPCv2}         & ResNet-$161$             & $305$M & $71.5$ & $90.1$\\
         \moco \cite{MOCO}          & ResNet-$50$ ($4\times$)  & $375$M & $68.6$ & -\\
         \simclr \cite{SimCLR}      & ResNet-$50$ ($4\times$)  & $375$M & $76.5$ & $93.2$\\
         \algo (ours)               & ResNet-$50$ ($4\times$)  & $375$M & $\cellcolor{pearDark!20}78.6$ &\cellcolor{pearDark!20} $94.2$\\
         \algo (ours)               & ResNet-$200$ ($2\times$) & $250$M & $\cellcolor{pearDark!20}\bf{79.6}$ &\cellcolor{pearDark!20} $\bf{94.8}$\\
         \bottomrule
    \end{tabular}
     \caption{\label{table:linear_classif_2}Other ResNet encoder architectures.}
\end{subtable}
\caption{\label{table:linear_classif}Top-1 and top-5 accuracies (in \%) under linear evaluation on ImageNet.}
\vspace{-0.5em}
\end{table}

\vfill
\paragraph{Semi-supervised training on ImageNet} 
Next, we evaluate the performance obtained when fine-tuning \algo's representation on a classification task with a small subset of ImageNet's \texttt{train\xspace} set, this time using label information. We follow the semi-supervised protocol of~\cite{kornblith2019better,dmlab,SimCLR,CPCv2} detailed in \Cref{app:semi_sup_details}, and use the same fixed splits of respectively $1\%$ and $10\%$ of ImageNet labeled training data as in~\cite{SimCLR}. We report both top-$1$ and top-$5$ accuracies on the  \texttt{test\xspace} set in \Cref{table:few_labels}. \algo consistently outperforms previous approaches across a wide range of architectures. 
Additionally, as detailed in \Cref{app:semi_sup_details}, \algo reaches $77.7\%$ top-$1$ accuracy with ResNet-50 when fine-tuning over $100\%$ of ImageNet labels. 

\vfill

\begin{table}[h]
\setlength\tabcolsep{3.5pt}
\begin{subtable}{.4\linewidth}
    \small
    \centering
    \begin{tabular}{l r r r r}
\toprule
    Method     & \multicolumn{2}{c}{Top-$1$} & \multicolumn{2}{c}{Top-$5$} \\
              &  $1\%$ & $10\%$            & $1\%$ & $10\%$ \\
    \midrule
    Supervised~\cite{zhai2019s4l} & $25.4$ & $56.4$ & $48.4$ & $80.4$ \\
    \midrule
    InstDisc                 & - & -                     & $39.2$ & $77.4$\\
    PIRL \cite{PIRL}         & - & -                     & $57.2$ & $83.8$\\
    \simclr \cite{SimCLR}    & $48.3$ & $65.6$           & $75.5$ & $87.8$\\
    \algo (ours)             & \cellcolor{pearDark!20} $\bf{53.2}$ & \cellcolor{pearDark!20} $\bf{68.8}$ &\cellcolor{pearDark!20} $\bf{78.4}$ & \cellcolor{pearDark!20} $\bf{89.0}$\\
    \bottomrule
\end{tabular}
\vspace{0.5em}
\caption{ResNet-50 encoder. }
\end{subtable}
\hfill
\begin{subtable}{.6\linewidth}
    \small
    \centering
    \begin{tabular}{l l r r r r r}
    \toprule
         Method     & Architecture & Param. & \multicolumn{2}{c}{Top-$1$} & \multicolumn{2}{c}{Top-$5$} \\
                    &              &         & $1\%$ & $10\%$            & $1\%$ & $10\%$ \\
         \midrule
         \CPCC \cite{CPCv2}       & ResNet-$161$            & $305$M     & - & - & $77.9$ & $91.2$\\
         \simclr \cite{SimCLR}    & ResNet-$50$ ($2\times$) & $94$M         & $58.5$ & $71.7$ & ${83.0}$ & $91.2$\\
          \algo (ours)            & ResNet-$50$ ($2\times$) & $94$M         & \cellcolor{pearDark!20} ${62.2}$ & \cellcolor{pearDark!20} ${73.5}$ &\cellcolor{pearDark!20} ${84.1}$              & \cellcolor{pearDark!20} ${91.7}$\\
         \simclr \cite{SimCLR}    & ResNet-$50$ ($4\times$) & $375$M     & $63.0$ & $74.4$ & $85.8$ & ${92.6}$\\
         \algo (ours)             & ResNet-$50$ ($4\times$) & $375$M     &\cellcolor{pearDark!20} ${69.1}$& \cellcolor{pearDark!20} ${75.7}$ &\cellcolor{pearDark!20} ${87.9}$ &\cellcolor{pearDark!20} $92.5$\\
         \algo (ours)             & ResNet-$200$ ($2\times$) & $250$M    &\cellcolor{pearDark!20} $\bf{71.2}$ & $\cellcolor{pearDark!20}\bf{77.7}$ &\cellcolor{pearDark!20} $\bf{89.5}$ &\cellcolor{pearDark!20} $\bf{93.7}$\\
         \bottomrule
    \end{tabular}
    \vspace{0.5em}
\caption{Other ResNet encoder architectures.}
\end{subtable}
\caption{\label{table:few_labels}Semi-supervised training with a fraction of ImageNet labels.}

\vspace{-0.8em}
\end{table}

\paragraph{Transfer to other classification tasks} We evaluate our representation on other classification datasets to assess whether the features learned on ImageNet (IN) are generic and thus useful across image domains, or if they are ImageNet-specific.
We perform linear evaluation and fine-tuning on the same set of classification tasks used in~\cite{SimCLR,kornblith2019better}, and carefully follow their evaluation protocol, as detailed in \Cref{app:transfer_details}. Performance is reported using standard metrics for each benchmark, and results are provided on a held-out \texttt{test\xspace} set after hyperparameter selection on a validation set. We report results in \Cref{tab:transfer_learning}, both for linear evaluation and fine-tuning. \algo outperforms \simclr on all benchmarks and the \supervisedbaseline~baseline on $7$ of the $12$ benchmarks, providing only slightly worse performance on the $5$ remaining benchmarks. \algo's representation can be transferred over to small images, e.g., CIFAR~\cite{cifar}, landscapes, e.g., SUN397~\cite{sun397} or VOC2007~\cite{pascal}, and textures, e.g., DTD~\cite{dtd}.

\begin{table}[t]
    \centering
    \scriptsize
    \setlength\tabcolsep{3.8pt}
\hspace{-0.1cm}
\begin{tabular}{l c c c c c c c c c c c c}
\cmidrule[\heavyrulewidth]{1-13}
{ Method} & { Food101} & { CIFAR10} & { CIFAR100} & { Birdsnap} & { SUN397}  & { Cars} & { Aircraft} & { VOC2007} & { DTD} & { Pets} & { Caltech-101} & { Flowers} \\
\midrule
\multicolumn{13}{l}{\emph{Linear evaluation:}}\\
\midrule
\algo (ours)&\cellcolor{pearDark!20}$\bf{75.3}$&\cellcolor{pearDark!20}$91.3$&\cellcolor{pearDark!20}$\bf{78.4}$&\cellcolor{pearDark!20}$\bf{57.2}$&\cellcolor{pearDark!20}$\bf{62.2}$&\cellcolor{pearDark!20}$\bf{67.8}$&\cellcolor{pearDark!20}$60.6$&\cellcolor{pearDark!20}$82.5$&\cellcolor{pearDark!20}$75.5$&\cellcolor{pearDark!20}$90.4$&\cellcolor{pearDark!20}$94.2$&\cellcolor{pearDark!20} $\bf{96.1}$\\
\simclr (repro)&$72.8$&$90.5$&$74.4$&$42.4$&$60.6$&$49.3$& $49.8$ & $81.4$&$\bf{75.7}$&$84.6$&$89.3$&$92.6$\\\simclr\cite{SimCLR}&$68.4$&$90.6$&$71.6$&$37.4$&$58.8$&$50.3$ &$50.3$&$80.5$&$74.5$&$83.6$&$90.3$&$91.2$\\\supervisedbaseline~\cite{SimCLR}&$72.3$&$\bf{93.6}$&$78.3$&$53.7$&$61.9$& $66.7$&$\bf{61.0}$&$\bf{82.8}$&$74.9$&$\bf{91.5}$&$\bf{94.5}$&$94.7$\\
\midrule
\multicolumn{13}{l}{\emph{Fine-tuned:}}\\
\midrule
\algo (ours) &\cellcolor{pearDark!20}$\bf{88.5}$&\cellcolor{pearDark!20}$\bf{97.8}$&\cellcolor{pearDark!20}$86.1$&\cellcolor{pearDark!20}$\bf{76.3}$&\cellcolor{pearDark!20}$63.7$&\cellcolor{pearDark!20}$91.6$&\cellcolor{pearDark!20}$\bf{88.1}$&\cellcolor{pearDark!20} $\bf{85.4}$ &\cellcolor{pearDark!20}$\bf{76.2}$&\cellcolor{pearDark!20}$91.7$&\cellcolor{pearDark!20}$\bf{93.8}$&\cellcolor{pearDark!20}$97.0$\\
\simclr (repro) & $87.5$ & $97.4$ & $85.3$ & $75.0$ & $63.9$ & $91.4$ & $87.6$ & $84.5$ & $75.4$ & $89.4$ & $91.7$ & $96.6$  \\
\simclr \cite{SimCLR}  & $88.2$ & $97.7$ & $85.9$ & $75.9$ & $63.5$ & $91.3$ & $88.1$ & $84.1$ & $73.2$ & $89.2$ & $92.1$ & $97.0$  \\
\supervisedbaseline~\cite{SimCLR} & $88.3$ & $97.5$ & $\bf{86.4}$ & $75.8$ & $\bf{64.3}$ & $\bf{92.1}$ & $86.0$ & $85.0$ & $74.6$ & $\bf{92.1}$ & $93.3$ & $\bf{97.6}$ \\
Random init \cite{SimCLR} & $86.9$ & $95.9$ & $80.2$ & $76.1$  & $53.6$ & $91.4$ & $85.9$ & $67.3$ & $64.8$ & $81.5$ & $72.6$ & $92.0$  \\
\bottomrule
\end{tabular}
     \vspace{0.5em}
     \caption{Transfer learning results from ImageNet (IN) with the standard ResNet-50 architecture.}
     \label{tab:transfer_learning}
     \vspace{-0.8em}
\end{table}

\paragraph{Transfer to other vision tasks} We evaluate our representation on different tasks relevant to computer vision practitioners, namely semantic segmentation, object detection and depth estimation. With this evaluation, we assess whether \algo's representation generalizes beyond classification tasks.

We first evaluate \algo on the VOC2012 semantic segmentation task as detailed in~\Cref{app:semantic_seg}, where the goal is to classify each pixel in the image~\cite{long2015fully}. We report the results in \Cref{tab:detect_segment}. \algo outperforms both the \supervisedbaseline~baseline ($+1.9$ mIoU) and \simclr ($+1.1$ mIoU).

Similarly, we evaluate on object detection by reproducing the setup in~\cite{MOCO} using a Faster R-CNN architecture~\cite{ren2015faster}, as detailed in~\Cref{app:object_det}. We fine-tune on  \texttt{trainval2007\xspace} and report results on  \texttt{test2007\xspace} using the standard AP$_{50}$ metric; \algo is significantly better than the \supervisedbaseline~baseline ($+3.1$ AP$_{50}$) and \simclr ($+2.3$ AP$_{50}$).

Finally, we evaluate on depth estimation on the NYU v2 dataset, where the depth map of a scene is estimated given a single RGB image. Depth prediction measures how well a network represents geometry, and how well that information can be localized to pixel accuracy~\cite{doersch2017multi}. The setup is based on~\cite{laina2016depth} and detailed in \Cref{app:depth}. We evaluate on the commonly used  \texttt{test\xspace} subset of $654$ images and report results using several common metrics in \Cref{table:depth}: relative (rel) error, root mean squared (rms) error, and the percent of pixels (pct) where the error, $\max (d_{gt}/d_{p}, d_{p}/d_{gt})$, is below $1.25^{n}$ thresholds where $d_p$ is the predicted depth and $d_{gt}$ is the ground truth depth~\cite{doersch2017multi}. \algo is better or on par with other methods for each metric. For instance, the challenging pct.\,$\!\,\!<\!\!1.25$ measure is respectively improved by $+3.5$ points and $+1.3$ points compared to supervised and \simclr baselines.

\vspace{0.5em}
\begin{table}[h]
\begin{subtable}{.35\linewidth}
\small
\begin{center}
\centering\hspace*{-1.cm}
\begin{tabular}{l r r }
\toprule
    Method     & AP$_{50}$ & mIoU \\
    \midrule
    \supervisedbaseline~\cite{MOCO} & $74.4$ & $74.4$  \\
    \midrule
    MoCo~\cite{MOCO}       & $74.9$ & $72.5$\\
    \simclr (repro)  & $75.2$ & $75.2$\\
    \algo (ours)   & \cellcolor{pearDark!20} $\bf{77.5}$  & \cellcolor{pearDark!20} $\bf{76.3}$\\
    \bottomrule
\end{tabular}
\caption{\label{tab:detect_segment}Transfer results in semantic\\  segmentation and object detection.}
\end{center}
\end{subtable}
\begin{subtable}{.65\linewidth}
\small
\centering
\vspace{-0.330cm} \hspace*{-0.7cm}
\begin{tabular}{l c c c c c}
\toprule
               & \multicolumn{3}{c}{Higher better} & \multicolumn{2}{c}{Lower better}\\
    {Method}&{pct.\,$\!<\!1.25$}&{pct.\,$\!<\!1.25^2$}&{pct.\,$\!<\!1.25^3$}&{rms} &{rel}\\
    \midrule
    \supervisedbaseline~\cite{laina2016depth}&$81.1$&$95.3$&$98.8$&$0.573$&$\bf{0.127}$\\
    \midrule
    \simclr (repro) &$83.3$&$96.5$&$99.1$&$0.557$&$0.134$\\
    \algo (ours) &\cellcolor{pearDark!20}$\bf{84.6}$&\cellcolor{pearDark!20}$\bf{96.7}$& \cellcolor{pearDark!20} $\bf{99.1}$&\cellcolor{pearDark!20}$\bf{0.541}$&\cellcolor{pearDark!20}$0.129$\vspace*{-0.05cm}\\
    \bottomrule
\end{tabular}
\caption{\label{table:depth}Transfer results on NYU v2 depth estimation.}
\end{subtable}
\caption{\label{tab:other_transfer}Results on transferring \algo's representation to other vision tasks.}
\vspace{-0.8em}
\end{table}

\section{Building intuitions with ablations}
\label{sec:ablation}
\label{sec:contrast}
We present ablations on \algo to give an intuition of its behavior and performance. For reproducibility, we run each configuration of parameters over three seeds, and report the average performance. We also report the half difference between the best and worst runs when it is larger than $0.25$. Although previous works perform ablations at $100$ epochs~\cite{SimCLR,Tian2020WhatMF}, we notice that relative improvements at $100$ epochs do not always hold over longer training. For this reason, we run ablations over $300$ epochs on $64$ TPU v$3$ cores, which yields consistent results compared to our baseline training of $1000$ epochs. For all the experiments in this section, we set the initial learning rate to $0.3$ with batch size $4096$, the weight decay to $10^{-6}$ as in \simclr~\cite{SimCLR} and the base target decay rate $\tau_\text{base}$ to $0.99$. In this section we report results in top-$1$ accuracy on ImageNet under the linear evaluation protocol as in \Cref{app:linear_eval_details}.

\paragraph{Batch size}
Among contrastive methods, the ones that draw negative examples from the minibatch suffer  performance drops when their batch size is reduced. \algo does not use negative examples and we expect it to be 
more robust to smaller batch sizes. To empirically verify this hypothesis, we train both \algo and \simclr using different batch sizes from $128$ to $4096$. To avoid re-tuning other hyperparameters, we average gradients over $N$ consecutive steps before updating the online network when reducing the batch size by a factor $N$. The target network is updated once every $N$ steps, after the update of the online network; we accumulate the $N$-steps in parallel in our runs.

\begin{figure}[h]
    \centering
    \begin{subfigure}{0.48\linewidth}
        \centering
        \vskip 0.1em
        \includegraphics[width=0.85\linewidth,bb=0 0 895 812]{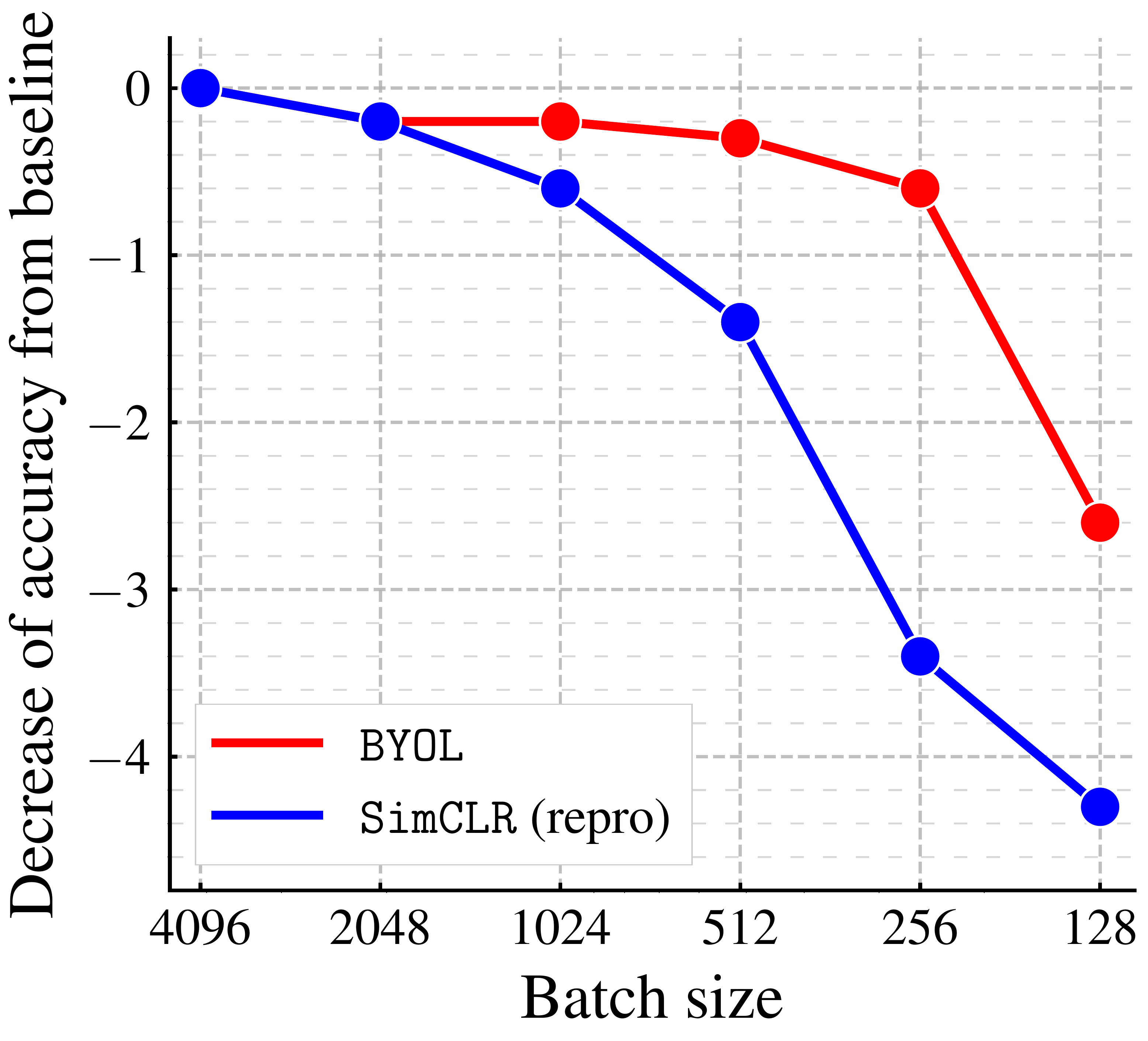}
        \caption{Impact of batch size}
        \label{fig:batchsize}
    \end{subfigure}%
    \hfill
    \begin{subfigure}{0.48\linewidth}
        \centering
        \includegraphics[width=0.85\linewidth,bb=0 0 926 853]{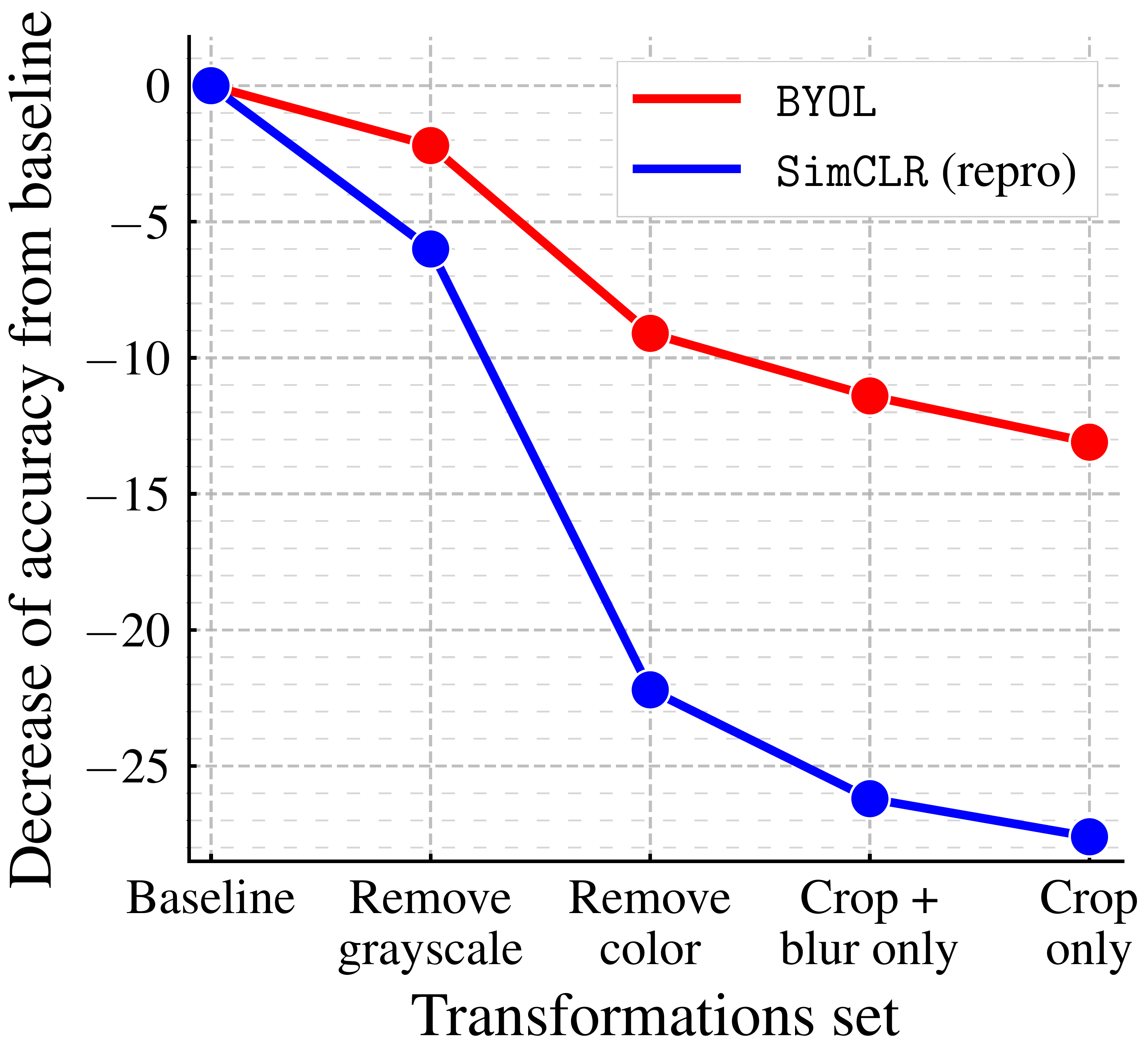}
        \caption{Impact of progressively removing transformations}
        \label{fig:transfo}
    \end{subfigure}
    \captionsetup{width=.95\linewidth}
    \caption{\label{fig:ablations}Decrease in top-$1$ accuracy (in \% points) of \algo and our own reproduction of \simclr at $300$ epochs, under linear evaluation on ImageNet.}
    \vspace{-0.7em}
\end{figure}

As shown in \Cref{fig:batchsize}, the performance of \simclr rapidly deteriorates with batch size, likely due to the decrease in the number of negative examples. In contrast, the performance of \algo remains stable over a wide range of batch sizes from $256$ to $4096$, and only drops for smaller values due to batch normalization layers in the encoder.\footnote{The only dependency on batch size in our training pipeline sits within the batch normalization layers.}

\paragraph{\Imageaugmentation s}
\label{subsec:robustness_tranfo}
Contrastive methods are sensitive to the choice of \imageaugmentation s. For instance, \simclr does not work well when removing color distortion from its \imageaugmentation s. As an explanation, \simclr shows that crops of the same image mostly share their color histograms. At the same time, color histograms vary across images. 
Therefore, when a contrastive task only relies on random crops as \imageaugmentation s, it can be mostly solved by focusing on color histograms alone. As a result the representation is not incentivized to retain information beyond color histograms.
To prevent that, \simclr adds color distortion to its set of \imageaugmentation s. Instead, \algo is incentivized to keep any information captured by the target representation into its online network, to improve its predictions. Therefore, even if \augview s of a same image share the same color histogram, \algo is still incentivized to retain additional features in its representation. For that reason, we believe that \algo is more robust to the choice of \imageaugmentation s than contrastive methods.

Results presented in \Cref{fig:transfo} support this hypothesis: the performance of \algo is much less affected than the performance of \simclr when removing color distortions from the set of \imageaugmentation s ($-9.1$ accuracy points for \algo, $-22.2$ accuracy points for \simclr). 
When \imageaugmentation s are reduced to mere random crops, \algo still displays good performance ($59.4\%$, \emph{i.e.} $-13.1$ points from $72.5\%$ ), while \simclr loses more than a third of its performance ($40.3\%$, \emph{i.e.}  $-27.6$ points from $67.9\%$). We report additional ablations in \Cref{app:imageaugment}.

\paragraph{Bootstrapping}

\algo uses the projected representation of a target network, whose weights are an exponential moving average of the weights of the online network, as target for its predictions. This way, the weights of the target network represent a delayed and more stable version of the weights of the online network. When the target decay rate is $1$, the target network is never updated, and remains at a constant value corresponding to its initialization. When the target decay rate is $0$, the target network is instantaneously updated to the online network at each step. 
There is a trade-off between updating the targets too often and updating them too slowly, as illustrated in \Cref{tab:ablation_bootstrap}. 
Instantaneously updating the target network ($\tau = 0$) destabilizes training, yielding very poor performance while never updating the target ($\tau = 1$) makes the training stable but prevents iterative improvement, ending with low-quality final representation. All values of the decay rate between $0.9$ and $0.999$ yield performance above $68.4\%$ top-$1$ accuracy at $300$ epochs. 

\newcommand{\misovspace}{\vspace{0.685em}}
\begin{table}[h]
\begin{subtable}{.49\linewidth}
    \small
    \centering
    \begin{tabular}{l r r}
            \toprule
              Target  & $\tau_\text{base}$ & Top-$1$  \\
                \midrule
                Constant random network & $1$ & $18.8 {\scriptstyle \pm 0.7}$ \\ 
                Moving average of online  & $0.999$ & $69.8$ \\
                Moving average of online  & $0.99$  & $\bf{72.5}$ \\
                Moving average of online  & $0.9$   & $68.4$ \\
                Stop gradient of online$^{\color{red}\dagger}$  & $0$ & $0.3$ \\
            \bottomrule
    \end{tabular}
    \caption{Results for different target modes.~$^\dagger$In the \emph{stop gradient of online}, $\tau = \tau_\text{base} = 0$ is kept constant throughout training. }
    \label{tab:ablation_bootstrap}
\end{subtable}
\hfill
\begin{subtable}{.49\linewidth}
    \centering
    \small
    \begin{tabular}{l c c c r}
    \toprule
    Method & Predictor & Target network & $\beta$ & Top-$1$  \\
    \midrule
    \algo    & \checkmark & \checkmark &  0 & $\bf{72.5}$ \\
      $-$    & \checkmark & \checkmark & 1 & $70.9$ \\
      $-$    &            & \checkmark & 1 & $70.7$\\
    \simclr  &            &            & 1 & $69.4$\\
      $-$    & \checkmark &            & 1 & $69.1$\\
      $-$    & \checkmark &            & 0 & $0.3$  \\
      $-$    &            & \checkmark & 0 & $0.2$  \\
      $-$    &            &            & 0 & $0.1$   \\
    \bottomrule
    \end{tabular}
     \caption{Intermediate variants between \algo and \simclr. }
     \label{tab:simclr_to_pbl}
\end{subtable}
\caption{Ablations  with top-$1$ accuracy (in \%) at $300$ epochs under linear evaluation on ImageNet.}
\label{table:ablation_bs_and_pbl_to_simclr}
\vspace{-0.8em}
\end{table}

\paragraph{Ablation to contrastive methods}
\label{subsec:simclr_to_pbl}
In this subsection, we recast \simclr and \algo using the same formalism to better understand  where the improvement of \algo over \simclr comes from. Let us consider the following objective that extends the InfoNCE objective \cite{CPC, poole2019variational} (see \Cref{app:neg-examples}),

\begin{align}
\text{InfoNCE}^{\alpha, \beta}_{\netparams} \eqdef
\label{eq:infonce}
 \frac{2}{B}\sum_{i=1}^B S_\netparams(v_i, v'_i)\!- \beta \cdot \frac{2\alpha}{B}\sum_{i=1}^B \ln\pa{\sum\limits_{j \neq i} \exp \frac{S_\netparams(v_i, v_j)}{\alpha} + \sum\limits_{j} \exp \frac{S_\netparams(v_i, v'_j)}{\alpha}}\CommaBin
\end{align}

where $\alpha>0$ is a fixed temperature, $\beta \in [0, 1]$ a weighting coefficient, $B$ the batch size, $v$ and $v'$ are batches of \augview s where for any batch index $i$, $v_i$ and $v'_i$ are \augview s from the same image; the real-valued function $S_\netparams$ quantifies pairwise similarity between \augview s. For any \augview~$u$ we denote $z_\netparams(u) \triangleq f_\netparams(g_\netparams(u))$ and $z_\targetparams(u) \triangleq f_\targetparams(g_\targetparams(u))$. For given $\phi$ and~$\psi$, we consider the normalized dot product

\begin{equation}
    S_\netparams(u_1, u_2) \eqdef \frac{\langle \phi(u_1), \psi(u_2) \rangle }{\|\phi(u_1)\|_2\cdot \|\psi(u_2)\|_2}\cdot
\end{equation}

Up to minor details (cf.\,\Cref{app:simclr_baseline}), we recover the \simclr loss with $\phi(u_1) = z_\netparams(u_1)$ (no predictor), $\psi(u_2) = z_\netparams(u_2)$ (no target network) and $\beta=1$. We recover the \algo loss when using a predictor and a target network, \emph{i.e.,} $\phi(u_1) = p_\netparams\pa{z_\netparams(u_1)}$ and $\psi(u_2) = z_\targetparams(u_2)$ with $\beta=0$. To evaluate the influence of the target network, the predictor and the coefficient~$\beta$, we perform an ablation over them. Results are presented in \Cref{tab:simclr_to_pbl} and more details are given in \Cref{app:neg-examples}.
 
The only variant that performs well without negative examples (i.e., with $\beta = 0$) is \algo, using \textit{both} a bootstrap target network \textit{and} a predictor. 
Adding the negative pairs to \algo's loss without re-tuning the temperature parameter hurts its performance. 
In \Cref{app:neg-examples}, we show that we can add back negative pairs and still match the performance of \algo with proper tuning of the temperature.

Simply adding a target network to \simclr already improves performance ($+1.6$ points). This sheds new light on the use of the target network in \moco~\cite{MOCO}, where the target network is used to provide more negative examples. Here, we show that by mere stabilization effect, even when using the same number of negative examples, using a target network is beneficial. 
Finally, we observe that modifying the architecture of $S_\netparams$ to include a predictor only mildly affects the performance of \simclr.

\paragraph{Network hyperparameters} In Appendix~\ref{sec:additional_ablation_results}, we explore how other network parameters may impact \algo's performance.
We iterate over multiple weight decays, learning rates, and projector/encoder architectures to observe that small hyperparameter changes do not drastically alter the final score. We note that removing the weight decay in either \algo or \simclr leads to network divergence, emphasizing the need for weight regularization in the self-supervised setting. Furthermore, we observe that changing the scaling factor in the network initialization~\cite{he2015delving} did not impact the performance (higher than $72\%$ top-$1$ accuracy).

\paragraph{Relationship with \meanteacher}\label{sec:mean_teacher} Another  semi-supervised approach, \meanteacher (MT)~\cite{tarvainen2017mean},  complements a supervised loss on few labels with an additional consistency loss. In~\cite{tarvainen2017mean}, this consistency loss is the $\ell_2$ distance between the logits from a \textit{student} network, and those of a temporally averaged version of the student network, called \textit{teacher}. 
Removing the predictor in \algo results in an unsupervised version of MT with no classification loss that uses image augmentations instead of the original architectural noise (e.g., dropout).
This variant of \algo collapses (Row 7 of~\Cref{table:ablation_bs_and_pbl_to_simclr}) which suggests that the additional predictor is critical to prevent collapse in an unsupervised scenario.

\paragraph{Importance of a near-optimal predictor}
\Cref{tab:simclr_to_pbl} already shows the importance of combining a predictor and a target network: the representation does collapse when either is removed. 
We further found that we can remove the target network without collapse by making the predictor near-optimal, either by (i) using an optimal \emph{linear} predictor (obtained by linear regression on the current batch) before back-propagating the error through the network ($52.5\%$ top-1 accuracy), or (ii) increasing the learning rate of the predictor ($66.5\%$ top-1). By contrast, increasing the learning rates of both projector \emph{and} predictor (without target network) yields poor results ($\approx 25\%$ top-1). See \Cref{app:importance_no_pred} for more details. 
This seems to indicate that keeping the predictor near-optimal at all times is important to preventing collapse, which may be one of the roles of \algo's target network. 

\section{Conclusion}
\label{sec:conclusion}
We introduced \algo, a new algorithm for self-supervised learning of image representations.
\algo learns its representation by predicting previous versions of its outputs, without using negative pairs. We show that \algo achieves state-of-the-art results on various benchmarks. In particular, under the linear evaluation protocol on ImageNet with a ResNet-$50$ ($1 \times$), \algo achieves a new state of the art and bridges most of the remaining gap between self-supervised methods and the supervised learning baseline of~\cite{SimCLR}. Using a ResNet-$200$~$(2\times)$, \algo reaches a top-$1$ accuracy of $79.6\%$  which improves over the previous state of the art ($76.8\%$) while using $30\%$ fewer parameters. 

Nevertheless, \algo remains dependent on existing sets of augmentations that are specific to vision applications. 
To generalize \algo to other modalities (e.g., audio, video, text, \dots) it is necessary to obtain similarly suitable augmentations for each of them. 
Designing such augmentations may require significant effort and expertise. 
Therefore, automating the search for these augmentations would be an important next step to generalize \algo to other modalities.

\clearpage

\section*{Broader impact}
The presented research should be categorized as research in the field of unsupervised learning. This work may inspire new algorithms, theoretical, and experimental investigation. The algorithm presented here can be used for many different vision applications and a particular use may have both positive or negative impacts, which is known as the dual use problem. Besides, as vision datasets could be biased, the representation learned by \algo could be susceptible to replicate these biases.

\section*{Acknowledgements}
The authors would like to thank the following people for their help throughout the process of writing this paper, in alphabetical order: Aaron van den Oord, Andrew Brock, Jason Ramapuram, Jeffrey De Fauw, Karen Simonyan, Katrina McKinney, Nathalie Beauguerlange, Olivier Henaff, Oriol Vinyals, Pauline Luc, Razvan Pascanu, Sander Dieleman, and the DeepMind team. We especially thank Jason Ramapuram and Jeffrey De Fauw, who provided the JAX SimCLR reproduction used throughout the paper.

\clearpage
\appendix
\section{Algorithm}
\label{app:algo}
\begin{algorithm}[!ht]
\DontPrintSemicolon
\SetAlgoLined
\SetKwInOut{Input}{Inputs}
\SetKwInOut{Output}{Output}
\Input{
\par
\begin{tabular}{l l}
$\mathcal{D}$, $\mathcal{T},$ and $\mathcal{T}^{\prime}$ & set of images and distributions of transformations\\
$\netparams$, $f_\netparams$, $g_\netparams,$ and $q_\netparams$ & initial online parameters, encoder, projector, and predictor\\
$\targetparams$, $f_\targetparams$, $g_\targetparams$ & initial target parameters, target encoder, and target projector\\
$\mathrm{optimizer}$ & optimizer, updates online parameters using the loss gradient\\
$K$ and $N$ & total number of optimization steps and batch size\\
$\{\tau_k\}_{k=1}^K$ and $\{\eta_k\}_{k=1}^K$ & target network update schedule and learning rate schedule\\
\end{tabular}}

\For{$k=1$ \KwTo $K$}{ 
$\mathcal{B} \gets \{x_i\sim\mathcal{D}\}_{i=1}^N$ \tcp*{sample a batch of $N$ images}
\For{$x_i\in\mathcal{B}$}{
$t \sim \mathcal{T} {\rm\ and\ } t^{\prime} \sim \mathcal{T}^{\prime}$ \tcp*{sample image transformations}
$z_1 \leftarrow g_\netparams\left(f_\netparams(t(x_i))\right)$ and $z_2 \leftarrow g_\netparams\left(f_\netparams(t^{\prime}(x_i))\right)$ \tcp*{compute projections}

$z^{\prime}_1 \leftarrow g_\targetparams\left(f_\targetparams(t^{\prime}(x_i))\right)$ and  $z^{\prime}_2 \leftarrow g_\targetparams\left(f_\targetparams(t(x_i))\right)$\tcp*{compute target projections}

$l_i \leftarrow  -2\cdot\pa{\frac{\langle q_\netparams(z_1), z'_1\rangle}{\normtwo{q_\netparams(z_1)}\cdot\normtwo{z'_1}} + \frac{\langle q_\netparams(z_2),  z'_2\rangle}{\normtwo{q_\netparams(z_2)}\cdot\normtwo{z'_2}}} $\tcp*{compute the loss for $x_i$}
}
$\dtheta \leftarrow  \frac{1}{N}\sum\limits_{i=1}^{N} \partial_\theta l_i$ \tcp*{compute the total loss gradient w.r.t.\,$\theta$}
$\netparams \leftarrow \mathrm{optimizer}(\netparams, \dtheta, \eta_k)$\tcp*{update online parameters}
$\targetparams \leftarrow \tau_k \targetparams + (1 - \tau_k) \netparams$\tcp*{update target parameters}
}

\Output{encoder $f_\netparams$}
\caption{\algo:  \longalgo}
\label{alg:PBL}
\end{algorithm}

\section{Image augmentations}
\label{app:transf}
During self-supervised training, \algo uses the following \imageaugmentation s (which are a subset of the ones presented in~\cite{SimCLR}):
\begin{itemize}
    \item random cropping: a random patch of the image is selected, with an area uniformly sampled between 8\% and 100\% of that of the original image, and an aspect ratio logarithmically sampled between $3/4$ and $4/3$. This patch is then resized to the target size of $224 \times 224$ using bicubic interpolation;
    \item optional left-right flip;
    \item color jittering: the brightness, contrast, saturation and hue of the image are shifted by a uniformly random offset applied on all the pixels of the same image. The order in which these shifts are performed is randomly selected for each patch;
    \item color dropping: an optional conversion to grayscale. When applied, output intensity for a pixel $(r, g, b)$ corresponds to its luma component, computed as $0.2989 r + 0.5870 g + 0.1140 b$;
    \item Gaussian blurring: for a $224 \times 224$ image, a square Gaussian kernel of size $23 \times 23$ is used, with a standard deviation uniformly sampled over $[0.1, 2.0]$;
    \item solarization: an optional color transformation $x \mapsto x \cdot \bOne_{\{x < 0.5\}} + (1 - x) \cdot \bOne_{\{x \ge 0.5\}}$ for pixels with values in $[0, 1]$.
\end{itemize}

Augmentations from the sets $\mathcal{T}$ and~$\mathcal{T}'$ (introduced in \Cref{sec:method}) are compositions of the above \imageaugmentation s in the listed order, each applied with a predetermined probability. The \imageaugmentation s parameters are listed in \Cref{tab:transformation_distributions}.

During evaluation, we use a center crop similar to~\cite{SimCLR}: images are resized to $256$ pixels along the shorter side using bicubic resampling, after which a $224 \times 224$ center crop is applied. In both training and evaluation, we normalize color channels by subtracting the average color and dividing by the standard deviation, computed on ImageNet, after applying the augmentations.

\begin{table}[ht]
    \small
    \centering
    \begin{tabular}{l l l} \toprule
        Parameter & $\mathcal{T}$ & $\mathcal{T}'$ \\ \midrule
        Random crop probability & $1.0$ & $1.0$ \\
        Flip probability & $0.5$ & $0.5$ \\
        Color jittering probability & $0.8$ & $0.8$ \\
        Brightness adjustment max intensity & $0.4$ & $0.4$ \\
        Contrast adjustment max intensity & $0.4$ & $0.4$ \\
        Saturation adjustment max intensity & $0.2$ & $0.2$ \\
        Hue adjustment max intensity & $0.1$ & $0.1$ \\
        Color dropping probability & $0.2$ & $0.2$ \\
        Gaussian blurring probability & $1.0$ & $0.1$ \\
        Solarization probability & $0.0$ & $0.2$ \\ \bottomrule
    \end{tabular}
     \vspace{0.5em}
    \caption{Parameters used to generate \imageaugmentation s.}
    \label{tab:transformation_distributions}
\end{table}

\section{Evaluation on ImageNet training}
\label{app:eval_details}

\subsection{Self-supervised learning evaluation on ImageNet}
\paragraph{Linear evaluation protocol on ImageNet}
\label{app:linear_eval_details}
As in~\cite{kolesnikov2019revisiting,kornblith2019better,SimCLR,MOCOv2}, we use the standard linear evaluation protocol on ImageNet, which consists in training a linear classifier on top of the frozen representation, \emph{i.e.}, without updating the network parameters nor the batch statistics. At training time, we apply spatial augmentations, i.e., random crops with resize to $224 \times 224$ pixels, and random flips. At test time, images are resized to $256$ pixels along the shorter side using bicubic resampling, after which a $224 \times 224$ center crop is applied. In both cases, we normalize the color channels by subtracting the average color and dividing by the standard deviation (computed on ImageNet), after applying the augmentations. We optimize the cross-entropy loss using \SGD with Nesterov momentum over $80$ epochs, using a batch size of $1024$ and a momentum of $0.9$. We do not use any regularization methods such as weight decay, gradient clipping~\cite{pascanu2013difficulty}, tclip~\cite{AMDIM}, or logits regularization. We finally sweep over $5$ learning rates $\{0.4, 0.3, 0.2, 0.1, 0.05\}$ on a local validation set (10009 images from ImageNet \texttt{train} set), and report the accuracy of the best validation hyperparameter on the \texttt{test} set (which is the public validation set of the original ILSVRC2012 ImageNet dataset).

\paragraph{Variant on linear evaluation on ImageNet}
In this paragraph only, we deviate from the protocol of~\cite{SimCLR,MOCOv2} and
propose another way of performing linear evaluation on top of a frozen representation. This method achieves better performance both in top-1 and top-5 accuracy. 

\begin{itemize}[leftmargin=*]
    \item We replace the spatial augmentations (random crops with resize to $224 \times 224$ pixels and random flips) with the pre-train augmentations of \Cref{app:transf}. This method was already used in~\cite{CPCv2} with a different subset of pre-train augmentations. 
    \item We regularize the linear classifier as in  \cite{AMDIM}\footnote{\url{https://github.com/Philip-Bachman/amdim-public/blob/master/costs.py}} by clipping the logits using a hyperbolic tangent function 
\[ {\rm tclip}(x) \triangleq \alpha \cdot \tanh( x / \alpha),\] where $\alpha$  is a positive scalar, and by adding a logit-regularization penalty term in the loss
\[
{\rm Loss}(x,y) \triangleq {\rm cross\_entropy}({\rm tclip}(x), y) + \beta \cdot {\rm average}({\rm tclip}(x)^2), 
\] 
where $x$ are the logits, $y$ are the target labels, and $\beta$ is the regularization parameter. We set $\alpha=20$ and $\beta=1e{-2}$.
\end{itemize}
 
We report in~\Cref{tab:recuit_enhanced} the top-1 and top-5 accuracy on ImageNet using this modified protocol. These modifications in the evaluation protocol increase the \algo's top-$1$ accuracy from $74.3\%$ to $74.8\%$ with a ResNet-$50$ ($1\times$).

\begin{table}[ht]
    \centering
    \small
    \begin{tabular}[ht]{l c c r r }
    \toprule
          Architecture      & Pre-train augmentations & Logits regularization &  Top-$1$ & Top-$5$         \\
         \midrule 
          \multirow{4}{*}{ResNet-$50$ ($1\times$)}  & & & $74.3$  & $91.6$ \\
          & \checkmark & & $74.4$& $91.8$ \\
          &  & \checkmark & $74.7$& $91.8$ \\ 
          &  \checkmark & \checkmark & $\bf{74.8}$& $\bf{91.8}$ \\ 
                    \midrule
          \multirow{4}{*}{ResNet-$50$ ($4\times$)}  & & & $78.6$  & $94.2$ \\
          & \checkmark & & $78.6$ & $94.3$ \\
          &  & \checkmark & $78.9$ & $94.3$ \\ 
          &  \checkmark & \checkmark & $\bf{79.0}$& $\bf{94.5}$ \\ 
          \midrule
          \multirow{4}{*}{ResNet-$200$ ($2\times$)} & &  & $79.6$  & $94.8$ \\
          & \checkmark & & $79.6$ & $94.8$ \\
          &  & \checkmark & $79.8$& $95.0$ \\ 
          &  \checkmark & \checkmark & $\bf{80.0}$& $\bf{95.0}$ \\ 
         \bottomrule
    \end{tabular}
    \vspace{0.5em}
\captionsetup{width=.9\linewidth}
\caption{\label{tab:recuit_enhanced}Different linear evaluation protocols on ResNet architectures by either replacing the spatial augmentations with pre-train augmentations, or regularizing the linear classifier. No pre-train augmentations and no logits regularization correspond to the evaluation protocol of the main paper, which is the same as in \cite{SimCLR,MOCOv2}. }
\end{table}  

\paragraph{Semi-supervised learning on ImageNet} 
\label{app:semi_sup_details}
We follow the semi-supervised learning protocol of~\cite{SimCLR,zhai2019s4l}. We first initialize the network with the parameters of the pretrained representation, and fine-tune it with a subset of ImageNet labels.
At training time, we apply spatial augmentations, i.e., random crops with resize to $224 \times 224$ pixels and random flips. At test time, images are resized to $256$ pixels along the shorter side using bicubic resampling, after which a $224 \times 224$ center crop is applied. In both cases, we normalize the color channels by subtracting the average color and dividing by the standard deviation (computed on ImageNet), after applying the augmentations. We optimize the cross-entropy loss using \SGD with Nesterov momentum. We used a batch size of $1024$, a momentum of $0.9$. We do not use any regularization methods such as weight decay, gradient clipping~\cite{pascanu2013difficulty}, tclip~\cite{AMDIM}, or logits rescaling. We sweep over the learning rate $\{0.01, 0.02, 0.05, 0.1, 0.005\}$ and the number of epochs $\{30, 50\}$ and select the hyperparameters achieving the best performance on our local validation set to report test performance.

\begin{figure}[ht]
    \centering
    \begin{subfigure}{0.48\linewidth}
        \centering
        \vskip 0.1em
        \includegraphics[width=\linewidth,bb = 0 0 872 791]{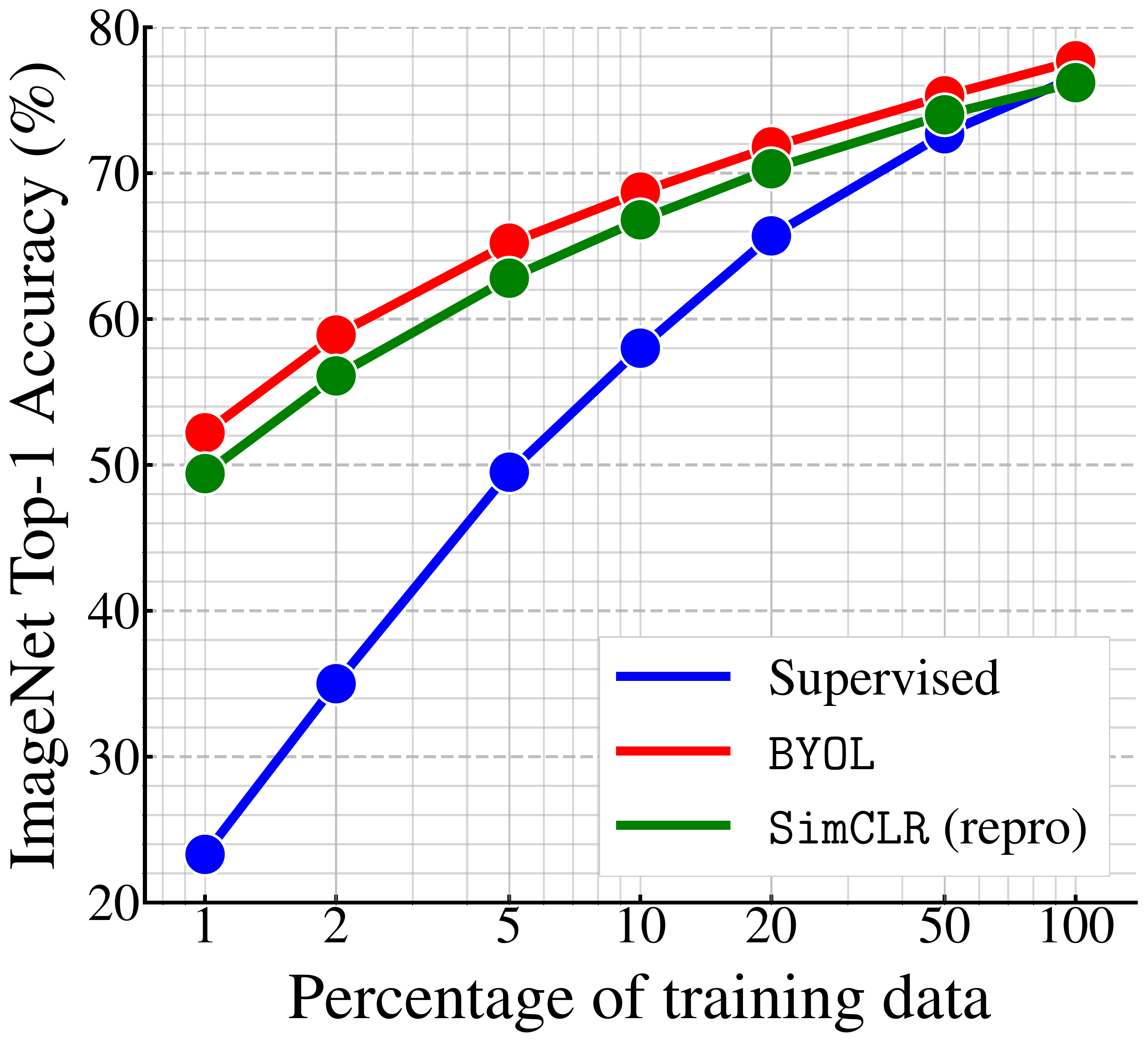}
        \caption{Top-1 accuracy}
        \label{fig:few_data_top1}
    \end{subfigure}
    \hfill
    \begin{subfigure}{0.48\linewidth}
        \centering
        \includegraphics[width=\linewidth,bb = 0 0 872 780]{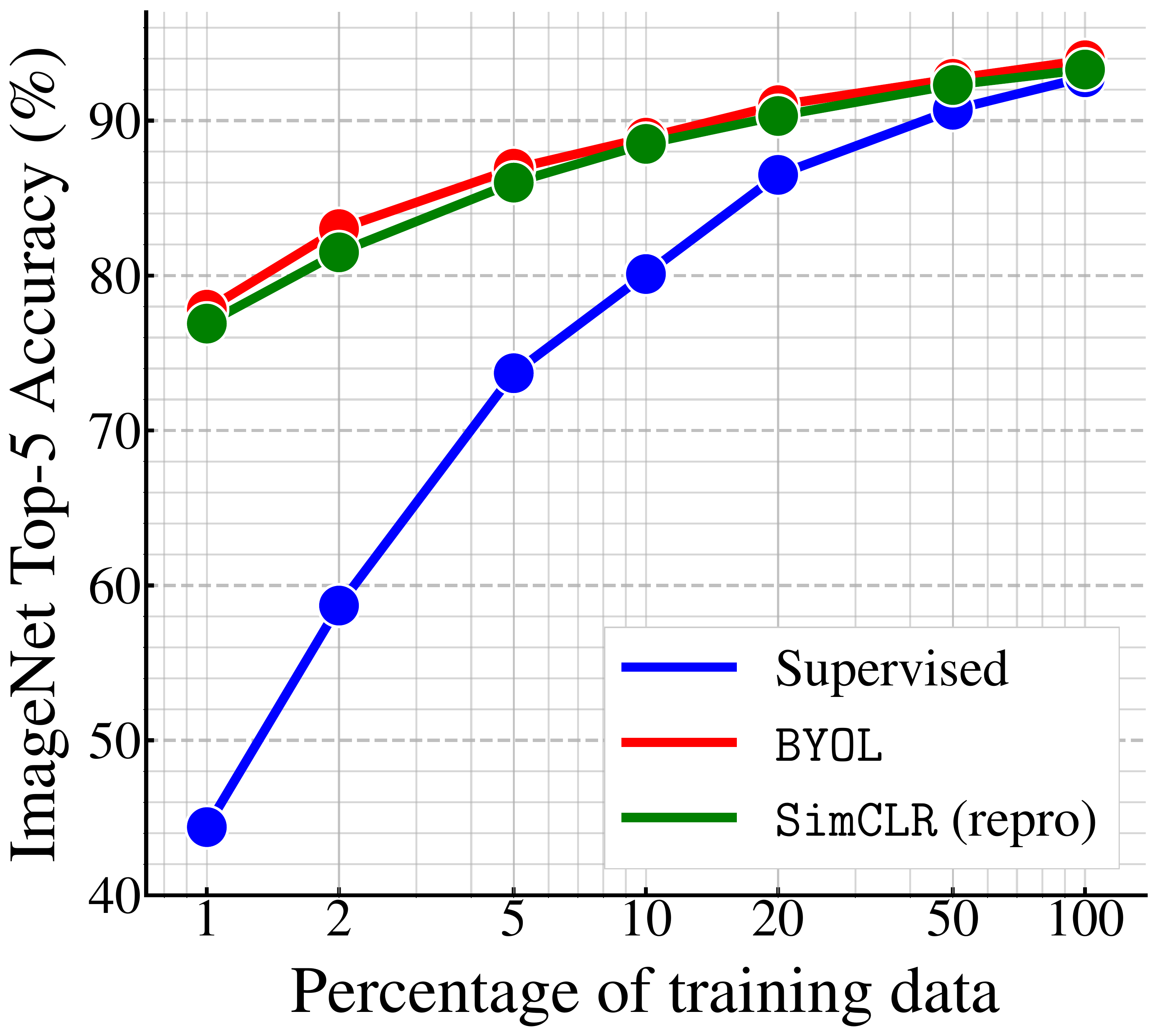}
        \caption{Top-5 accuracy}
        \label{fig:few_data_top5}
    \end{subfigure}
    \caption{\label{fig:few_data} Semi-supervised training with a fraction of ImageNet labels on a ResNet-$50$ ($\times 1$). }
\end{figure}

\begin{table}[ht]
    \centering
    \small
    \setlength\tabcolsep{9.3pt}
\begin{tabular}[t]{l r r l r r}
\toprule
\multicolumn{3}{l}{\emph{Supervised:}} & \multicolumn{3}{l}{\emph{Semi-supervised ($100\%$):}} \\
 Method      &    Top-$1$ & Top-$5$ &  Method    &    Top-$1$ & Top-$5$     \\
\midrule
Supervised\cite{SimCLR}                  & $76.5$ & $-$ & \simclr~\cite{SimCLR}     & $76.0$ & $93.1$ \\
AutoAugment~\cite{cubuk2019randaugment}  & $77.6$ & $93.8$ & \simclr (repro)       & $76.5$ & $93.5$ \\
MaxUp~\cite{gong2020maxup}               & $\bf{78.9}$ & $\bf{94.2}$ & \algo & \cellcolor{pearDark!20} $\bf{77.7}$ & \cellcolor{pearDark!20} $\bf{93.9}$ \\ 
\bottomrule
\end{tabular}
    \vspace{0.5em}
    \captionsetup{width=.7\linewidth}
    \caption{Semi-supervised training with the full ImageNet on a ResNet-$50$ ($\times 1$). We also report other fully supervised methods for extensive comparisons.}
    \label{tab:finetune100}
\end{table}

In \Cref{table:few_labels} presented in the main text, we fine-tune the representation over the $1$\% and $10$\% ImageNet splits from \cite{SimCLR} with various ResNet architectures. 

\begin{figure}[ht]
    \centering
    \includegraphics[width=\linewidth,trim={6cm 4cm 6cm 6cm},clip, bb=0 0 2160 1440]{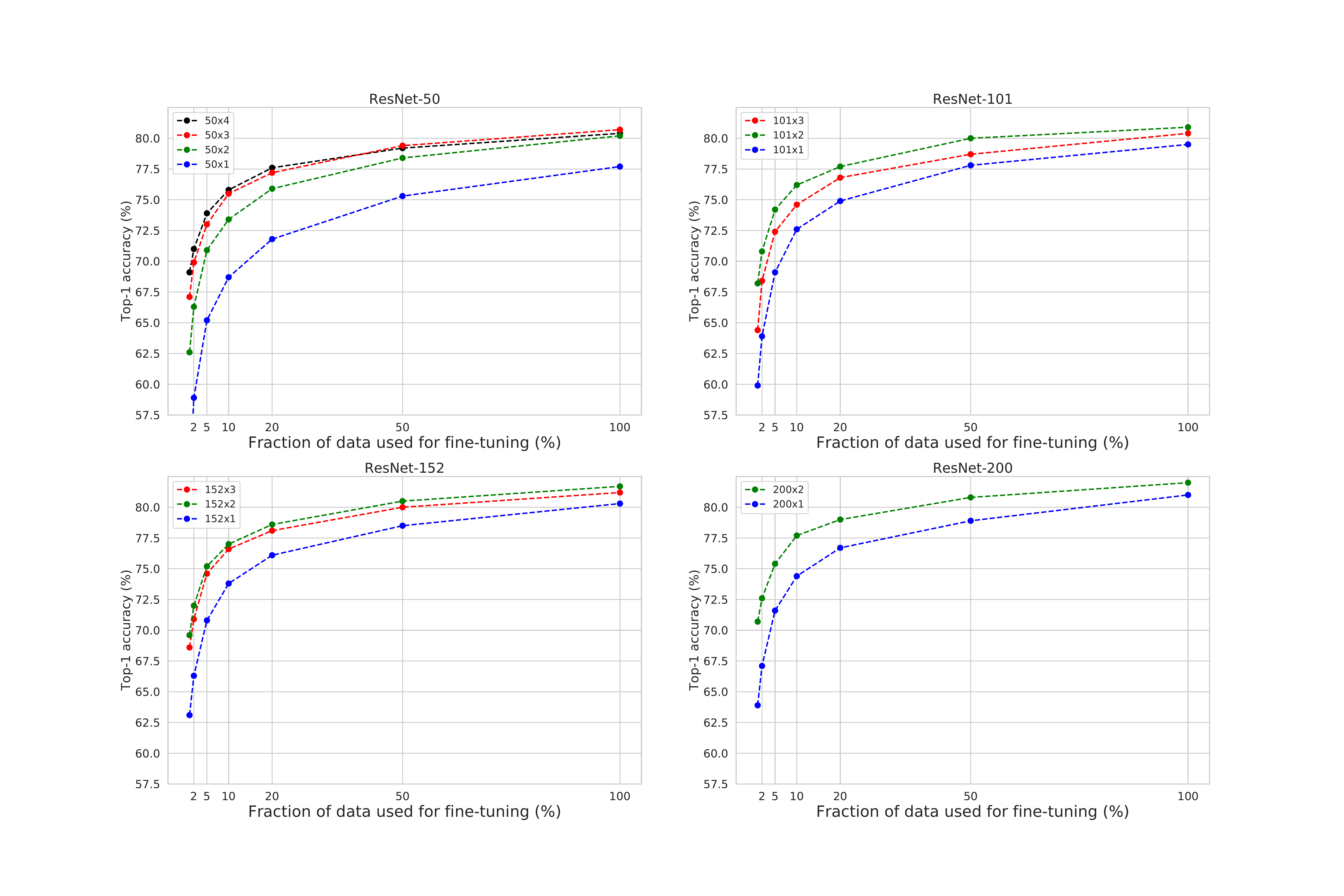}
    \caption{\label{fig:few_data_all_resnet}  Semi-supervised training with a fraction of ImageNet labels on multiple ResNets architecture pretrained with \algo. Note that large networks are facing overfitting problems.}
\end{figure}

In \Cref{fig:few_data}, we fine-tune the representation over $1$\%, $2$\%, $5$\%, $10$\%, $20$\%, $50$\%, and $100$\% of the ImageNet dataset as in~\cite{CPCv2} with a ResNet-$50$ ($1\times$) architecture, and compare them with a supervised baseline and a fine-tuned \simclr representation. 
In this case and contrary to~\Cref{table:few_labels} we don't reuse the splits from \simclr but we create our own via a balanced selection. 
In this setting, we observed that tuning a \algo representation always outperforms a supervised baseline trained from scratch. In \Cref{fig:few_data_all_resnet}, we then fine-tune the representation over multiple ResNet architectures. We observe that the largest networks are prone to overfitting as they are outperformed by ResNets with identical depth but smaller scaling factor. This overfitting is further confirmed when looking at the training and evaluation loss: large networks have lower training losses, but higher validation losses than some of their slimmer counterparts. Regularization methods are thus recommended when tuning on large architectures.

Finally, we fine-tune the representation over the full ImageNet dataset. We report the results in \Cref{tab:finetune100} along with supervised baselines trained on ImageNet. We observe that fine-tuning the SimCLR checkpoint does not yield better results (in our reproduction, which matches the results reported in the original paper~\cite{SimCLR}) than using a random initialization ($76.5$ top-$1$). Instead, \algo's initialization checkpoint leads to a high final score ($77.7$ top-$1$), higher than the vanilla supervised baseline of \cite{SimCLR}, matching the strong supervised baseline of AutoAugment\cite{cubuk2019randaugment} but still $1.2$ points below the stronger supervised baseline~\cite{gong2020maxup}, which uses advanced supervised learning techniques.

\subsection{Linear evaluation on larger architectures and supervised baselines}
\label{app:larger_architectures}

\begin{table}[ht]
    \centering
    \small
    \begin{tabular}{lrrrrrrc}
\toprule
 &&& \multicolumn{2}{c}{\algo} & \multicolumn{2}{c}{Supervised (ours)} & Supervised~\cite{SimCLR} \\
 Architecture  & Multiplier  &   Weights &   Top-1 &   Top-5 & Top-1 & Top-5 & Top-1 \\
\midrule
 ResNet-50 &$1 \times$   &            24M &            $74.3$ &           $91.6$ & $76.4$ & $92.9$ & $76.5$\\
 ResNet-101 &$1 \times$  &            43M &            $76.4$ &           $93.0$ & $78.0$ & $94.0$ & - \\
 ResNet-152 &$1 \times$  &            58M &            $77.3$ &           $93.7$ & $79.1$ & $94.5$ & - \\
 ResNet-200 &$1 \times$  &            63M &            $77.8$ &           $93.9$ & $79.3$ & $94.6$ & - \\
 ResNet-50 &$2 \times$   &            94M &            $77.4$ &           $93.6$ & $79.9$ & $95.0$ & $77.8$ \\
 ResNet-101 &$2 \times$  &           170M &            $78.7$ &           $94.3$ & $80.3$ & $95.0$ & - \\
 ResNet-50 &$3 \times$   &           211M &            $78.2$ &           $93.9$ & $80.2$ & $95.0$ & - \\
 ResNet-152 &$2 \times$  &           232M &            $79.0$ &           $94.6$ & $80.6$ & $95.3$ & - \\
 ResNet-200 &$2 \times$  &           250M &            $\bf{79.6}$ &      $\bf{94.9}$ & $80.1$ & $95.2$ & - \\
 ResNet-50 &$4 \times$   &           375M &            $78.6$  &         $94.2$ & $80.7$ & $\bf{95.3}$ & $78.9$ \\
 ResNet-101 &$3 \times$  &           382M &            $78.4$ &          $94.2$ & $80.7$ & $\bf{95.3}$ & - \\
 ResNet-152 &$3 \times$  &           522M &            $79.5$ &          $94.6$ & $\bf{80.9}$ & $95.2$ & - \\
\bottomrule
\end{tabular}
     \vspace{0.5em}
     \captionsetup{width=.7\linewidth}
     \caption{Linear evaluation of \algo on ImageNet using larger encoders. Top-$1$ and top-$5$ accuracies are reported in \%.}
     \label{tab:large_resnet_perf}
\end{table}

\begin{figure}[ht]
    \centering
    \vskip 0.1em
    \includegraphics[width=0.95\linewidth,bb=0 0 1149 995]{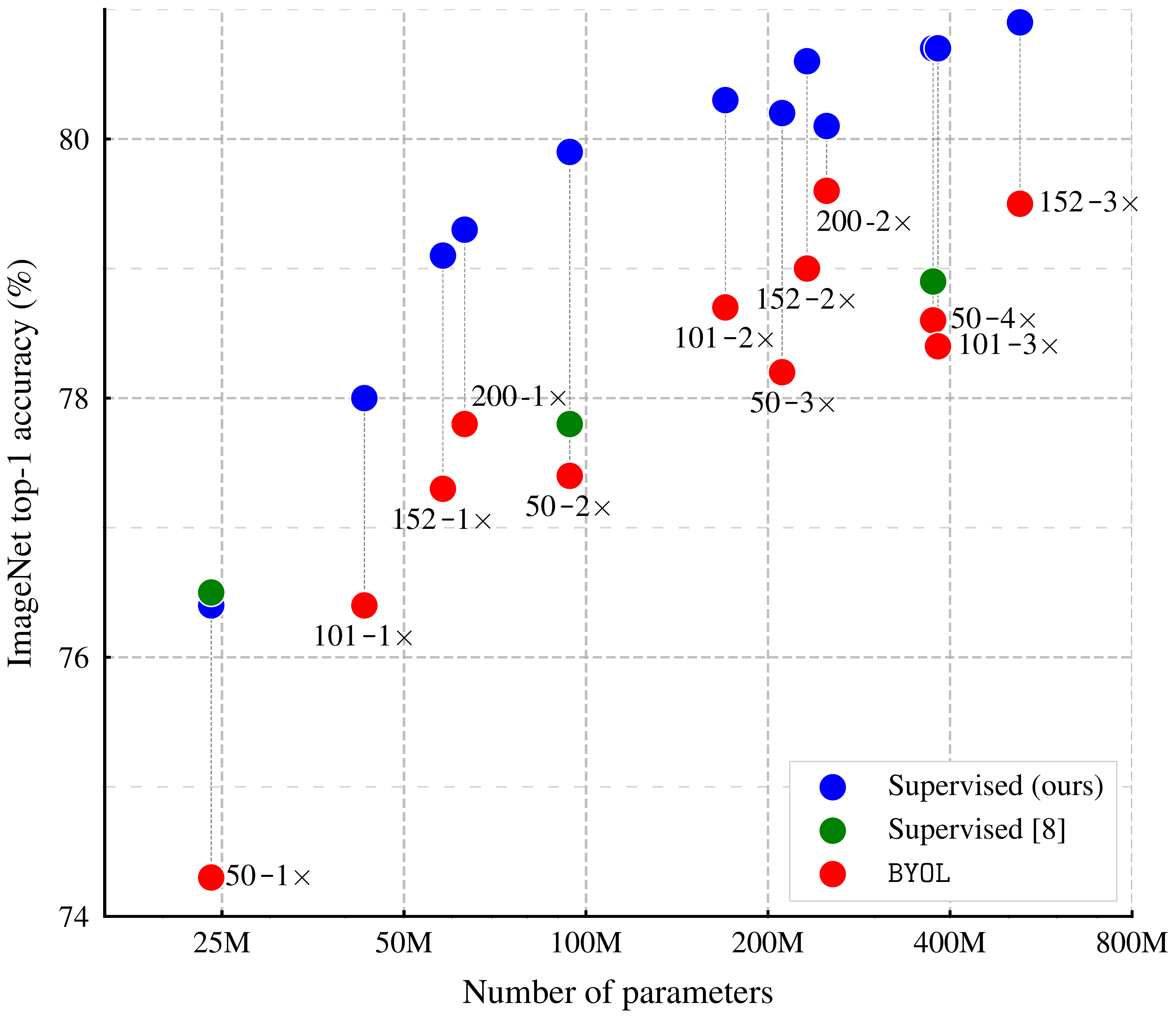}
    \captionsetup{width=.9\linewidth}
    \caption{Results for linear evaluation of \algo compared to fully supervised baselines with various ResNet architectures. Our supervised baselines are ran with RandAugment~\cite{cubuk2019randaugment} augmentations.}
    \label{fig:byol_architectures}
\end{figure}

Here we investigate the performance of \algo with deeper and wider ResNet architectures. We compare ourselves to the best supervised baselines from~\cite{SimCLR} when available (rightmost column in \cref{tab:large_resnet_perf}), which are also presented in~\Cref{fig:holygrail}. Importantly, we close in on those baselines using the ResNet-$50$ ($2 \times$) and the ResNet-$50$ ($4\times$) architectures, where we are within $0.4$ accuracy points of the supervised performance. To the best of our knowledge, this is the first time that the gap to supervised has been closed to such an extent using a self-supervised method under the linear evaluation protocol.
Therefore, in order to ensure fair comparison, and suspecting that the supervised baselines' performance in~\cite{SimCLR} could be even further improved with appropriate data augmentations, we also report on our own reproduction of strong supervised baselines. We use RandAugment~\cite{cubuk2019randaugment} data augmentation for all large ResNet architectures (which are all version $1$, as per~\cite{ResNet}). We train our supervised baselines for up to $200$ epochs, using SGD with a Nesterov momentum value of $0.9$, a cosine-annealed learning rate after a $5$ epochs linear warmup period, weight decay with a value of $1e-4$, and a label smoothing~\cite{szegedy2016rethinking} value of $0.1$. Results are presented in \Cref{fig:byol_architectures}.

\clearpage

\section{Transfer to other datasets}
\label{app:transfer_details}

\subsection{Datasets}
\label{app:datasets}
\begin{table}[ht]
   \small
    \centering
    \resizebox{\textwidth}{!}{
    \begin{tabular}{l r r r r r r r}
\toprule
Dataset & Classes & Original train examples & Train examples & Valid.\,examples & Test examples & Accuracy measure & Test provided\\
\midrule
ImageNet~\cite{ILSVRC15} &  $1000$ & $1281167$ & $1271158$ & $10009$ & $50000$ & Top-1 accuracy & -\\
Food101~\cite{food101}  &$101$ &$75750$ & $68175$ &$7575$ & $25250$ & Top-1 accuracy & - \\
CIFAR-10~\cite{cifar}  &$10$ &$50000$ & $45000$ & $5000$ &$10000$ & Top-1 accuracy & - \\
CIFAR-100~\cite{cifar} &$100$ &$50000$ & $44933$ & $5067$ &$10000$ & Top-1 accuracy & - \\
Birdsnap~\cite{Birdsnap} & $500$ & $47386$ & $42405$ &  $4981$ & $2443$ & Top-1 accuracy & - \\
Sun397 (split 1)~\cite{sun397}  &$397$ &$19850$ & $15880$ &$3970$ &$19850$ & Top-1 accuracy & - \\
Cars~\cite{carsdataset} & $196$ & $8144$ & $6494$ & $1650$ & $8041$ & Top-1 accuracy & - \\
Aircraft~\cite{aircraft} & $100$ & $3334$ & $3334$ & $3333$ & $3333$ & Mean per-class accuracy & Yes \\
PASCAL-VOC2007 ~\cite{pascal} &$20$ &$5011$ & $2501$ &$2510$ &$4952$ & 11-point mAP / AP50 & - \\
PASCAL-VOC2012 ~\cite{pascal} &$21$ &$10582$ & $-$ &$2119$ &$1449$ & Mean IoU & - \\
DTD (split 1) ~\cite{dtd} & $47$ & $1880$ & $1880$ & $1880$ & $1880$ & Top-1 accuracy & Yes \\
Pets~\cite{pets} & $37$ & $3680$ & $2940$ & $740$ & $3669$ & Mean per-class accuracy& - \\
Caltech-101~\cite{caltech101} & $101$ & $3060$ & $2550$ & $510$ & $6084$ & Mean per-class accuracy & - \\
Places365~\cite{zhou2017places} & $365$ & $1803460$ & $1803460$ & $-$ & $36500$ & Top-1 accuracy & - \\
Flowers~\cite{flowers} &$102$ & $1020$ & $1020$ & $1020$ & $6149$ & Mean per-class accuracy & Yes \\
\bottomrule
\end{tabular}
    }
     \vspace{0.5em}
     \captionsetup{width=.9\linewidth}
     \caption{Characteristics of image datasets used in transfer learning. When an official test split with labels is not publicly available, we use the official validation split as test set, and create a held-out validation set from the training examples.}
     \label{tab:datasets_transfer}
\end{table}
We perform transfer via linear classification and fine-tuning on the same set of datasets as in~\cite{SimCLR}, namely  Food-$101$ dataset~\cite{food101}, CIFAR-$10$~\cite{cifar}
and CIFAR-$100$~\cite{cifar}, Birdsnap~\cite{Birdsnap}, the SUN$397$ scene dataset~\cite{sun397}, Stanford Cars~\cite{carsdataset}, FGVC Aircraft~\cite{aircraft}, the PASCAL VOC $2007$ classification
task~\cite{pascal}, the Describable Textures Dataset (DTD)~\cite{dtd}, Oxford-IIIT Pets~\cite{pets}, Caltech-$101$~\cite{caltech101}, and Oxford $102$ Flowers~\cite{flowers}. As in~\cite{SimCLR}, we used the validation sets specified by the dataset creators to select hyperparameters for FGVC Aircraft, PASCAL VOC
2007, DTD, and Oxford 102 Flowers. On other datasets, we use the validation examples as test set, and hold out a subset of the training examples that we use as validation set. We use standard metrics for each datasets:
\begin{itemize}
    \item \textit{Top-$1$}: We compute the proportion of correctly classified examples.
    \item \textit{Mean per class}: We compute the top-$1$ accuracy for each class separately and then compute the empirical mean over the classes.
    \item \textit{Point 11-mAP}: We compute the empirical mean \emph{average precision} as defined in~\cite{pascal}.
    \item \textit{Mean IoU:} We compute the empirical mean Intersection-Over-Union as defined in~\cite{pascal}.
    \item \textit{AP50:} We compute the Average Precision as defined in~\cite{pascal}.
\end{itemize}

We detail the validation procedures for some specific datasets: 
\begin{itemize}
    \item For Sun397~\cite{sun397}, the original dataset specifies 10 train/test splits, all of which contain 50 examples/images of 397 different classes. We use the first train/test split. The original dataset specifies no validation split and therefore, the training images have been further subdivided into 40 images per class for the train split and 10 images per class for the valid split.
    \item For Birdsnap~\cite{Birdsnap}, we use a random selection of valid images with the same number of images per category as the test split.
    \item For DTD~\cite{dtd}, the original dataset specifies 10 train/validation/test splits, we only use the first split.
    \item For Caltech-101~\cite{caltech101}, the original does not dataset specifies any train/test splits. We have followed the approach used in~\cite{donahue2014decaf}: This file defines datasets for 5 random splits of 25 training images per category, with 5 validation images per category and the remaining images used for testing.
    \item For ImageNet, we took the last $10009$ last images of the official tensorflow ImageNet split.
    \item For Oxford-IIIT Pets, the valid set consists of 20 randomly selected images per class.	
\end{itemize}
Information about the dataset are summarized in \Cref{tab:datasets_transfer}.

\subsection{Transfer via linear classification}
We follow the linear evaluation protocol of~\cite{kolesnikov2019revisiting,kornblith2019better,SimCLR} that we detail next for completeness. 
We train a regularized multinomial logistic regression classifier on top of the frozen representation, i.e., with frozen pretrained parameters and without re-computing batch-normalization statistics.
In training and testing, we do not perform any image augmentations; images are resized to $224$ pixels along the shorter side using bicubic resampling and then normalized with ImageNet statistics.
Finally, we minimize the cross-entropy objective using \LBFGS with $\ell_2$-regularization, where we select the regularization parameters from a range of $45$ logarithmically-spaced values between $10^{-6}$ and $10^5$. After choosing the best-performing hyperparameters on the validation set, the model is retrained on combined training and validation images together, using the chosen parameters. The final accuracy is reported on the test set.

\subsection{Transfer via fine-tuning}
We follow the same fine-tuning protocol as in~\cite{CPCv2,kolesnikov2019revisiting,dmlab,SimCLR} that we also detail for completeness. Specifically, we initialize the network with the parameters of the pretrained representation. At training time, we apply spatial transformation, i.e., random crops with resize to $224 \times 224$ pixels and random flips. At test time, images are resized to $256$ pixels along the shorter side using bicubic resampling, after which a $224 \times 224$ center crop is extracted. In both cases, we normalize the color channels by subtracting the average color and dividing by the standard deviation (computed on ImageNet), after applying the augmentations.
We optimize the loss using \SGD with Nesterov momentum for $20 000$ steps with a batch size of $256$ and with a momentum of $0.9$. We set the momentum parameter for the batch normalization statistics to $\max(1- 10 / s, 0.9)$ where $s$ is the number of steps per epoch. The learning rate and weight decay are selected respectively with a grid of seven logarithmically spaced learning rates between $0.0001$ and $0.1$, and $7$ logarithmically-spaced values of weight decay between $10^{-6}$ and $10^{-3}$, as well as no weight decay. These values of weight decay are divided by the learning rate. After choosing the best-performing hyperparameters on the validation set, the model is retrained on combined training and validation images together, using the chosen parameters. The final accuracy is reported on the test set.

\subsection{Implementation details for semantic segmentation}
\label{app:semantic_seg}
We use the same fully-convolutional network (FCN)-based~\cite{long2015fully} architecture as~\cite{MOCO}. The backbone consists of the convolutional layers in ResNet-$50$. The $3\times 3$ convolutions in the conv$5$ blocks use dilation $2$ and stride $1$. This is followed by two extra $3\times 3$ convolutions with $256$ channels, each followed by batch normalization and ReLU activations, and a $1\times 1$ convolution for per-pixel classification. The dilation is set to $6$ in the two extra $3\times 3$ convolutions. The total stride is $16$ (FCN-$16$s \cite{long2015fully}).

We train on the \texttt{train\textunderscore aug2012\xspace} set and report results on \texttt{val2012\xspace}. Hyperparameters are selected on a $2119$ images held-out validation set. We use a standard per-pixel softmax cross-entropy loss to train the FCN. Training is done with random scaling (by a ratio in $\left[0.5, 2.0\right]$), cropping, and horizontal flipping. The crop size is $513$. Inference is performed on the $\left[513, 513\right]$ central crop. For training we use a batch size of $16$ and weight decay of $0.0001$. We select the base learning rate by sweeping across $5$ logarithmically spaced values between $10^{-3}$ and $10^{-1}$. The learning rate is multiplied by $0.1$ at the $70$-th and $90$-th percentile of training. We train for $30000$ iterations, and average the results on 5 seeds.

\subsection{Implementation details for object detection }
\label{app:object_det}

For object detection, we follow prior work on Pascal detection transfer~\cite{doersch2017multi,doersch2015unsupervised} wherever possible.  We use a Faster R-CNN~\cite{ren2015faster} detector with a R$50$-C$4$ backbone with a frozen representation. The R$50$-C$4$ backbone ends with the conv$4$ stage of a ResNet-$50$, and the box prediction head consists of the conv$5$ stage (including global pooling).
We preprocess the images by applying multi-scale augmentation (rescaling the image so its longest edge is between $480$ and $1024$ pixels) but no other augmentation. We use an asynchronous \texttt{SGD} optimizer with $9$ workers and train for $1.5$M steps. We used an initial learning rate of $10^{-3}$, which is reduced to $10^{-4}$ at 1M steps and to $10^{-5}$ at $1.2$M steps.

\subsection{Implementation details for depth estimation }
\label{app:depth}
For depth estimation, we follow the same protocol as in~\cite{laina2016depth}, and report its core components for completeness.
We use a standard ResNet-$50$ backbone and feed the conv$5$ features into $4$ fast up-projection blocks with respective filter sizes $512$, $256$, $128$, and $64$. We use a reverse Huber loss function for training~\cite{laina2016depth, owen2007robust}.

The original NYU Depth v$2$ frames of size $\left[640, 480\right]$ are down-sampled by a factor $0.5$ and center-cropped to $\left[304, 228\right]$ pixels. Input images are randomly horizontally flipped and the following color transformations are applied:
\begin{itemize}
\item \textit{Grayscale} with an application  probability of $0.3$.
\item \textit{Brightness} with a maximum brightness difference of $0.1255$.
\item \textit{Saturation} with a saturation factor randomly picked in the interval $\left[0.5, 1.5\right]$.
\item \textit{Hue} with a hue adjustment factor randomly picked in the interval $\left[-0.2, 0.2\right]$.
\end{itemize}
We train for $7500$ steps with batch size $256$, weight decay $0.0005,$ and learning rate $0.16$ (scaled linearly from the setup of \cite{laina2016depth} to account for the bigger batch size).

\subsection{Further comparisons on PASCAL and NYU v2 Depth}

For completeness, ~\Cref{tab:detect_segment_full} and~\ref{table:app_depth_full} extends~\Cref{tab:other_transfer} with other published baselines which use comparable networks.  We see that in almost all settings, \algo outperforms these baselines, even when those baselines use more data or deeper models.  
One notable exception is RMS error for NYU Depth prediction, which is a metric that's sensitive to outliers.
The reason for this is unclear, but one possibility is that the network is producing higher-variance predictions due to being more confident about a test-set scene's similarities with those in the training set.
\begin{table}[ht]
\begin{center}
\small
\centering
\begin{tabular}{l r r }
\toprule
    Method     & AP$_{50}$ & mIoU \\
    \midrule
    \supervisedbaseline~\cite{MOCO} & $74.4$ & $74.4$  \\
    \midrule
    RelPos~\cite{doersch2015unsupervised}, by~\cite{doersch2017multi}$^{*}$ & $66.8$ & - \\
    Multi-task~\cite{doersch2017multi}$^{*}$ & $70.5$ & - \\
    LocalAgg~\cite{zhuang2019local} & $69.1$ & - \\
    MoCo~\cite{MOCO} & $74.9$ & $72.5$\\
    MoCo + IG-1B~\cite{MOCO} & $75.6$ & $73.6$\\
    CPC\cite{CPCv2}$^{**}$ & $76.6$ & -\\
    \simclr (repro)  & $75.2$ & $75.2$\\
    \algo (ours)   & \cellcolor{pearDark!20} $\bf{77.5}$  & \cellcolor{pearDark!20} $\bf{76.3}$\\
    \bottomrule
\end{tabular}
\vspace{.2cm}
\caption{\label{tab:detect_segment_full}Transfer results in semantic segmentation and object detection.}\vspace*{-.2cm}
\end{center}
\caption*{$^{*}$ uses a larger model (ResNet-$101$). $^{**}$ uses an even larger model (ResNet-$161$).}
\end{table}

\begin{table}[ht]
\small
\begin{center}
\begin{tabular}{l c c c c c}
\toprule
               & \multicolumn{3}{c}{Higher better} & \multicolumn{2}{c}{Lower better}\\
    { Method}&{ pct.\,$\!<\!1.25$}&{ pct.\,$\!<\!1.25^2$}&{ pct.\,$\!<\!1.25^3$}&{ rms} &{ rel}\\
    \midrule
    \supervisedbaseline~\cite{laina2016depth}&$81.1$&$95.3$&$98.8$&$0.573$&$\bf{0.127}$\\
    \midrule
    RelPos~\cite{doersch2015unsupervised}, by~\cite{doersch2017multi}$^{*}$ & $80.6$ & $94.7$ & $98.3$ & $\bf{0.399}$ & $0.146$ \\
    Color~\cite{Zhang2016ColorfulIC}, by~\cite{doersch2017multi}$^{*}$ & $76.8$ & $93.5$ & $97.7$ & $0.444$ & $0.164$ \\
    Exemplar~\cite{dosovitskiy2014discriminative,doersch2017multi}$^{*}$ & $71.3$ & $90.6$ & $96.5$ & $0.513$ & $0.191$ \\
    Mot. Seg.~\cite{pathak2017learning}, by~\cite{doersch2017multi}$^{*}$ & $74.2$ & $92.4$ & $97.4$ & $0.473$ & $0.177$ \\
    Multi-task~\cite{doersch2017multi}$^{*}$ & $79.3$ & $94.2$ & $98.1$ & $0.422$ & $0.152$ \\
    \simclr (repro) &$83.3$&$96.5$&$99.1$&$0.557$&$0.134$\\
    \algo (ours) &\cellcolor{pearDark!20}$\bf{84.6}$&\cellcolor{pearDark!20}$\bf{96.7}$& \cellcolor{pearDark!20} $\bf{99.1}$&\cellcolor{pearDark!20}$0.541$&\cellcolor{pearDark!20}$0.129$\\
    \bottomrule
\end{tabular}\vspace*{.1cm}
\caption{\label{table:app_depth_full}Transfer results on NYU v2 depth estimation.}
\end{center}
\end{table}
\vspace{-1em}

\section{Pretraining on Places 365}
\label{app:train-places}
To ascertain that \algo learns good representations on other datasets, we applied our representation learning protocol on the scene recognition dataset Places$365$-Standard~\cite{zhou2017places} before performing linear evaluation. This dataset contains $1.80$ million training images and $36500$ validation images with labels, making it roughly similar to ImageNet in scale. We reuse the \emph{exact} same parameters as in \Cref{sec:experiments} and train the representation for $1000$ epochs, using \algo and our \simclr reproduction. Results for the linear evaluation setup (using the protocol of \Cref{app:linear_eval_details} for ImageNet and Places365, and that of \Cref{app:transfer_details} on other datasets) are reported in \Cref{tab:transfer_learning_from_places}.

Interestingly, the representation trained by using \algo on Places365 (\algo-PL) consistently outperforms that of \simclr on the same dataset, but underperforms the \algo representation trained on ImageNet (\algo-IN) on all tasks except Places365 and SUN397~\cite{sun397}, another scene understanding dataset. Interestingly, all three unsupervised representation learning methods achieve a relatively high performance on the Places365 task; for comparison, reference~\cite{zhou2017places} (in its linked repository) reports a top-1 accuracy of $55.2\%$ for a ResNet-$50$v2 trained from scratch using labels on this dataset.

\begin{table}[ht]
    \small
    \centering
     \setlength\tabcolsep{3.3pt}
\hspace{-0.1cm}
\begin{tabular}{l c c c c c c c c c c c c c}
\cmidrule[\heavyrulewidth]{1-14}
{\scriptsize Method} &  {\scriptsize Places365} & {\scriptsize ImageNet} & {\scriptsize Food101} & {\scriptsize CIFAR10} & {\scriptsize CIFAR100} & {\scriptsize Birdsnap} & {\scriptsize SUN397}  & {\scriptsize Cars} & {\scriptsize Aircraft} & {\scriptsize DTD} & {\scriptsize Pets} & {\scriptsize Caltech-101} & {\scriptsize Flowers} \\
\midrule
\algo-IN & $51.0$ & $74.3$ & $75.3$ & $91.3$ & $78.4$ & $57.3$ & $62.6$ & $67.2$ & $60.6$ & $76.5$ & $90.4$ & $94.3$ & $96.1$ \\ \midrule
\algo-PL & \cellcolor{pearDark!20}$\bf{53.2}$ & \cellcolor{pearDark!20}$\bf{58.5}$ & \cellcolor{pearDark!20}$\bf{64.7}$ & \cellcolor{pearDark!20}$\bf{84.5}$ & \cellcolor{pearDark!20}$\bf{66.1}$ & \cellcolor{pearDark!20}$\bf{28.8}$ & \cellcolor{pearDark!20}$\bf{64.2}$ & \cellcolor{pearDark!20}$\bf{55.6}$ & \cellcolor{pearDark!20}$\bf{55.9}$ & \cellcolor{pearDark!20}$\bf{68.5}$ & \cellcolor{pearDark!20}$\bf{66.1}$ & \cellcolor{pearDark!20}$\bf{84.3}$ & \cellcolor{pearDark!20}$\bf{90.0}$ \\
\simclr-PL & $53.0$ & $56.5$ & $61.7$ & $80.8$ & $61.1$ & $21.2$ & $62.5$ & $40.1$ & $44.3$ & $64.3$ & $59.4$ & $77.1$ & $85.9$ \\
\bottomrule
\end{tabular}
     \vspace{1em}
    \captionsetup{width=.9\linewidth} 
     \caption{Transfer learning results (linear evaluation, ResNet-$50$) from Places365 (PL). For comparison purposes, we also report the results from \algo trained on ImageNet (\algo-IN).}
     \label{tab:transfer_learning_from_places}
\vspace{-1.4em}
\end{table}

\section{Additional ablation results}\label{sec:additional_ablation_results}
To extend on the above results, we provide additional ablations obtained using the same experimental setup as in~\Cref{sec:ablation}, \emph{i.e.,} 300 epochs, averaged over $3$ seeds with the initial learning rate set to $0.3$, the batch size to $4096$, the weight decay to $10^{-6}$ and the base target decay rate $\tau_\text{base}$ to $0.99$ unless specified otherwise. Confidence intervals correspond to the half-difference between the maximum and minimum score of these seeds; we omit them for half-differences lower than $0.25$ accuracy points.

\subsection{Architecture settings}
\Cref{tab:archi-all-ablations} shows the influence of projector and predictor architecture on \algo. We examine the effect of different depths for both the projector and predictor, as well as the effect of the projection size. We do not apply a ReLU activation nor a batch normalization on the final linear layer of our MLPs such that a depth of $1$ corresponds to a linear layer.  Using the default projector and predictor of depth $2$ yields the best performance. 

\begin{table}[ht!]
\begin{minipage}{.45\linewidth}
\begin{subtable}{\linewidth}
    \small
    \begin{center}
        \begin{tabular}{P{1cm}P{1cm}rr}
        \toprule
           Proj.\,$g_\netparams$ \newline depth &   Pred.\,$q_\netparams$\newline depth & Top-$1$     & Top-$5$   \\
        \midrule
            \multirow{3}{*}{$1$}    & $1$   & $61.9$        & $86.0$ \\
                                    & $2$   & $65.0$        & $86.8$  \\
                                    & $3$   & $65.7$        & $86.8$  \\ \midrule
            \multirow{3}{*}{$2$}    & $1$   & $71.5$        & $90.7$ \\
                                    & $2$   & $\bf{72.5}$   & $\bf{90.8}$ \\
                                    & $3$   & $71.4$        & $90.4$  \\ \midrule
            \multirow{3}{*}{$3$}    & $1$   & $71.4$        & $90.4$ \\
                                    & $2$   & $72.1$        & $90.5$  \\
                                    & $3$   & $72.1$        & $90.5$  \\
        \bottomrule
        \end{tabular}
         \caption{Projector and predictor depth (i.e. the number of Linear layers). }
        \label{table:ablation-proj-depth}
    \end{center}
\end{subtable}
\end{minipage}
\phantom{aaaaa}
\begin{minipage}{.45\linewidth}
\begin{subtable}{\linewidth}
    \small
    \begin{center}
        \begin{tabular}{P{2cm}rr}
        \toprule
           Projector\,$g_\netparams$\newline output dim & Top-$1$ & Top-$5$ \\
        \midrule
                         $16$ & $69.9 {\scriptstyle \pm 0.3}$ & $89.9$ \\
                         $32$ & $71.3$ & $90.6$ \\
                         $64$ & $72.2$ & $90.9$ \\
                        $128$ & $72.5$ & $91.0$ \\
                        $256$ & $72.5$ & $90.8$ \\
                        $512$ & $\bf{72.6}$ & $\bf{91.0}$ \\
        \bottomrule
        \end{tabular}
         \caption{Projection dimension.}
         \label{table:ablation-proj-size}
     \end{center}
\end{subtable}
\end{minipage}
\caption{\label{tab:archi-all-ablations}Effect of architectural settings where top-$1$ and top-$5$ accuracies are reported in \%.}
\end{table}

\Cref{tab:ablation-lr} shows the influence of the initial learning rate on \algo. Note that the optimal value depends on the number of training epochs. \Cref{table:ablation-weight-decay} displays the influence of the weight decay on \algo. 

\begin{table}[ht!]
\hspace{-2em}
\begin{subtable}{.5\linewidth}
    \small
    \begin{center}
        \begin{tabular}{lrr} 
        \toprule
           Learning & &  \\
           rate & Top-$1$      & Top-$5$    \\
        \midrule
                 $0.01$ & $34.8 {\scriptstyle \pm 3.0}$ & $60.8 {\scriptstyle \pm3.2}$ \\
                 $0.1$  & $65.0 $ & $87.0$ \\
                 $0.2$  & $71.7$ & $90.6$ \\
                 $0.3$  & $\bf{72.5}$ & $\bf{90.8}$ \\
                 $0.4$  & $72.3$ & $90.6$ \\
                 $0.5$  & $71.5$ & $90.1$ \\
                 $1$    & $69.4$ & $89.2$ \\
        \bottomrule
        \end{tabular}
         \caption{Base learning rate.}
        \label{tab:ablation-lr}
    \end{center}
\end{subtable}
\phantom{aaaaa}
\begin{subtable}{0.5\linewidth}
    \small
    \begin{center}
        \begin{tabular}{P{2cm}rr}
        \toprule
           Weight decay \newline coefficient & Top-$1$ & Top-$5$ \\
        \midrule
              $1\cdot 10^{-7}$   & $72.1$ & $90.4$ \\
              $5\cdot 10^{-7}$   & $\bf{72.6}$ & $\bf{91.0}$ \\
              $1\cdot 10^{-6}$   & $72.5$ & $90.8$ \\
              $5\cdot 10^{-6}$   & $71.0 {\scriptstyle \pm 0.3}$ & $90.0$ \\
              $1\cdot 10^{-5}$   & $69.6 {\scriptstyle \pm 0.4}$ & $89.3$ \\
        \bottomrule
        \end{tabular}
         \caption{Weight decay.}
         \label{table:ablation-weight-decay}
     \end{center}
\end{subtable}
\caption{\label{tab:lr-ema-ablation}Effect of learning rate and weight decay. We note that \algo's performance is quite robust within a range of hyperparameters. We also observe that setting the weight decay to zero may lead to unstable results (as in \simclr).}
\end{table}

\subsection{Batch size}
We run a sweep over the batch size for both \algo and our reproduction of \simclr. As explained in~\Cref{sec:contrast}, when reducing the batch size by a factor $N$, we average gradients over $N$ consecutive steps and update the target network once every $N$ steps. 
We report in \Cref{tab:batchsize-all-ablations}, the performance of both our reproduction of \simclr and \algo for batch sizes between $4096$ (\algo and \simclr default) down to $64$. We observe that the performance of \simclr deteriorates faster than the one of \algo which stays mostly constant for batch sizes larger than $256$. We believe that the performance at batch size $256$ could match the performance of the large $4096$ batch size with proper parameter tuning when accumulating the gradient. We think that the drop in performance at batch size $64$ in \cref{tab:batchsize-all-ablations} is mainly related to the ill behaviour of batch normalization at low batch sizes~\cite{GroupNorm}.

\begin{table}[ht!]
    \centering\small
    \begin{tabular}{lrrrr}
    \cmidrule[\heavyrulewidth]{1-5}
     Batch &             \multicolumn{2}{c}{Top-$1$}    &  \multicolumn{2}{c}{Top-$5$}  \\
     size  &             \algo (ours) & \simclr (repro)    & \algo (ours) & \simclr (repro)  \\
    \midrule
    $4096$ & $\bf{72.5}$ & $67.9$ & $\bf{90.8}$ & $88.5$ \\
    $2048$ & $72.4$ & $67.8$ & $90.7$  & $88.5$ \\
    $1024$ & $72.2$ & $67.4$ & $90.7$  & $88.1$ \\
    $512$ & $72.2$  & $66.5$ & $90.8$  & $87.6$ \\
    $256$ & $71.8$  & $64.3 {\scriptstyle \pm 2.1}$& $90.7$  & $86.3 {\scriptstyle \pm1.0}$ \\
    $128$ & $69.6 {\scriptstyle \pm 0.5}$& $63.6$ & $89.6$ & $85.9$ \\
    $64$ & $59.7 {\scriptstyle \pm 1.5}$ & $59.2 {\scriptstyle \pm 2.9}$& $83.2 {\scriptstyle \pm1.2}$ & $83.0 {\scriptstyle \pm1.9}$ \\
    \bottomrule
    \end{tabular}
    \vspace{0.5em}
    \caption{\label{tab:batchsize-all-ablations}Influence of the batch size. }
\end{table}

\subsection{\Imageaugmentation s}
\label{app:imageaugment}
\Cref{tab:im-transfo-all-ablations} compares the impact of individual image transformations on \algo and \simclr. \algo is more resilient to changes of image augmentations across the board. For completeness, we also include an ablation with symmetric parameters across both views; for this ablation, we use a Gaussian blurring w.p.\,of $0.5$ and a solarization w.p.\,of $0.2$ for both $\mathcal T$ and $\mathcal T'$, and recover very similar results compared to our baseline choice of parameters.

\begin{table}[ht!]
    \centering\small
    \begin{tabular}{lrrrr}
    \cmidrule[\heavyrulewidth]{1-5}
                        &             \multicolumn{2}{c}{Top-$1$}    &  \multicolumn{2}{c}{Top-$5$}  \\
    Image augmentation  &             \algo (ours) & \simclr (repro)    & \algo (ours) & \simclr (repro)  \\
    \midrule
    Baseline &      $\bf{72.5}$ & $67.9$ & $\bf{90.8}$  & $88.5$ \\
    \midrule
    Remove flip  & $71.9$  & $67.3$ & $90.6$  & $88.2$ \\
    Remove blur  & $71.2$  & $65.2$ & $90.3$  & $86.6$ \\
    Remove color (jittering and grayscale) & $63.4 {\scriptstyle \pm 0.7}$  & $45.7$ & $85.3 {\scriptstyle \pm 0.5}$  & $70.6$ \\
    Remove color jittering & $71.8$  & $63.7$ & $90.7$  & $85.9$ \\
    Remove grayscale  & $70.3$  & $61.9$ & $89.8$  & $84.1$ \\
    Remove blur in $\mathcal T'$ & $72.4$ & $67.5$ & $90.8$ & $88.4$ \\
    Remove solarize in $\mathcal T'$ & $72.3$ & $67.7$ & $90.8$ & $88.2$ \\
    Remove blur and solarize in $\mathcal T'$ & $72.2$ & $67.4 $ & $90.8$  & $88.1$ \\
    Symmetric blurring/solarization & $72.5$ & $68.1$ & $90.8$ & $88.4$ \\
    Crop only & $59.4 {\scriptstyle \pm 0.3}$  & $40.3 {\scriptstyle \pm 0.3}$ & $82.4$  & $64.8 {\scriptstyle \pm 0.4}$ \\
    Crop and flip only & $60.1 {\scriptstyle \pm 0.3}$  & $40.2$ & $83.0 {\scriptstyle \pm 0.3}$  & $64.8$ \\
    Crop and color only & $70.7$ & $64.2$ & $90.0$ & $86.2$ \\
    Crop and blur only & $61.1 {\scriptstyle \pm 0.3}$  & $41.7$ & $83.9$  & $66.4$ \\
    \bottomrule
    \end{tabular}
    \vspace{0.5em}
    \caption{\label{tab:im-transfo-all-ablations}Ablation on image transformations.}
\end{table}

\begin{table}[ht!]
\centering \small
\begin{tabular}{rrrr}
\toprule
    Loss weight $\beta$ & Temperature $\alpha$ & Top-$1$  & Top-$5$     \\
\midrule
 $0$                    & $0.1$  & $72.5$ & $90.8$ \\
\midrule
 \multirow{7}{*}{$0.1$} & $0.01$ & $72.2$ & $90.7$ \\
                        & $0.1$  & $72.4$ & $90.9$ \\
                        & $0.3$  & $\bf{72.7}$ & $91.0$ \\
                        & $1$    & $72.6$ & $90.9$ \\
                        & $3$    & $72.5$ & $90.9$ \\
                        & $10$   & $72.5$ & $90.9$ \\
\midrule
\multirow{7}{*}{$0.5$}  & $0.01$ & $70.9$ & $90.2$ \\
                        & $0.1$  & $72.0$ & $90.8$ \\
                        & $0.3$  & $\bf{72.7}$ & $\bf{91.2}$ \\
                        &              $1$    & $72.7$ & $91.1$ \\
                        &              $3$    & $72.6$ & $91.1$ \\
                        &             $10$     & $72.5$ & $91.0$ \\
\midrule
\multirow{6}{*}{$1$}    & $0.01$ & $53.9 {\scriptstyle \pm 0.5}$ & $77.5 {\scriptstyle \pm0.5}$ \\
                        & $0.1$  & $70.9$ & $90.3$ \\
                        & $0.3$  & $\bf{72.7}$ & $91.1$ \\
                        &              $1$    & $\bf{72.7}$ & $91.1$ \\
                        &              $3$    & $72.6$ & $91.0$ \\
                        &             $10$    & $72.6$ & $91.1$ \\
\bottomrule
\end{tabular}\vspace{0.25cm}
\captionsetup{width=.9\linewidth}
\caption{\label{tab:app_neg_term_2}Top-1 accuracy in \% under linear evaluation protocol at 300 epochs of sweep over the temperature $\alpha$ and the dispersion term weight $\beta$ when using a predictor and a target network.}
\end{table}

\subsection{Details on the relation to contrastive methods}
\label{app:neg-examples}

As mentioned in \Cref{sec:contrast}, the \algo loss~\Cref{eq:cosine-loss} can be derived from the InfoNCE loss
\begin{equation}
\label{eq:infoNCE_beta}
\text{InfoNCE}^{\alpha, \beta}_{\netparams} \eqdef
 \frac{2}{B}\sum_{i=1}^B S_\netparams(v_i, v'_i)\!-\! \frac{2\alpha\cdot\beta}{B}\sum_{i=1}^B \ln\pa{\sum\limits_{j \neq i} \exp \frac{S_\netparams(v_i, v_j)}{\alpha} + \sum\limits_{j} \exp \frac{S_\netparams(v_i, v'_j)}{\alpha}}\CommaBin
\end{equation}
with
\begin{equation}
    S_\netparams(u_1, u_2) \eqdef \frac{\langle \phi(u_1), \psi(u_2) \rangle }{\|\phi(u_1)\|_2\cdot \|\psi(u_2)\|_2}\cdot
\end{equation}

The InfoNCE loss, introduced in \cite{CPC}, can be found in factored form in \cite{poole2019variational} as
\begin{equation}
\text{InfoNCE}_{\netparams} \eqdef
 \frac{1}{B}\sum_{i=1}^B \ln\cfrac{f(v_i, v'_i)}{\frac{1}{B}\sum\limits_{j} \exp f(v_i, v'_j)}\,\cdot
\end{equation}

As in \simclr~\cite{SimCLR} we also use negative examples given by $(v_i, v_j)_{j\neq i}$ to get
\begin{align}
 &\frac{1}{B}\sum_{i=1}^B \ln\cfrac{\exp f(v_i, v'_i)}{\frac{1}{B}\sum\limits_{j\neq i} \exp f(v_i, v_j) + \frac{1}{B}\sum\limits_{j} \exp f(v_i, v'_j)} \\
 &\qquad = \ln B + \frac{1}{B}\sum_{i=1}^Bf(v_i, v'_i) - \frac{1}{B}\sum_{i=1}^B \ln\pa{\sum\limits_{j\neq i} \exp f(v_i, v_j) + \sum\limits_{j} \exp f(v_i, v'_j)}.
 \label{eq:infoNCE_bis}
\end{align}

To obtain \Cref{eq:infoNCE_beta} from \Cref{eq:infoNCE_bis}, we subtract $\ln B$ (which is independent of $\netparams$), multiply by $2\alpha$, take $f(x,y) = S_\netparams(x,y)/\alpha$ and finally multiply the second (negative examples) term by $\beta$. Using $\beta=1$ and dividing by $2\alpha$ gets us back to the usual InfoNCE loss as used by \simclr.

In our ablation in \Cref{tab:simclr_to_pbl}, we set the temperature $\alpha$ to its best value in the \simclr setting (i.e., $\alpha = 0.1$). With this value, setting $\beta$ to 1 (which adds negative examples), in the \algo setting (i.e., with both a predictor and a target network) hurts the performances. 
In~\Cref{tab:app_neg_term_2}, we report results of a sweep over both the temperature $\alpha$ and the weight parameter
$\beta$ with a predictor and a target network where \algo corresponds to~$\beta = 0$. 
No run significantly outperforms \algo and some values of $\alpha$ and $\beta$ hurt the performance. While the best temperature for \simclr (without the target network and a predictor) is $0.1$, after adding a predictor and a target network the best temperature $\alpha$ is higher than~$0.3$.

Using a target network in the loss has two effects: stopping the gradient through the prediction targets and stabilizing the targets with averaging. Stopping the gradient through the target change the objective while averaging makes the target stable and stale. In~\Cref{tab:simclr_to_pbl} we only shows results of the ablation when either using the online network as the prediction target (and flowing the gradient through it) or with a target network (both stopping the gradient into the prediction targets and computing the prediction targets with a moving average of the online network). We shown in~\Cref{tab:simclr_to_pbl} that using a target network is beneficial but it has two distinct effects we would like to understand from which effect the improvement comes from. 
We report in \Cref{tab:app_neg_term_1} the results already in~\Cref{tab:simclr_to_pbl} but also when the prediction target is computed with a stop gradient of the online network (the gradient does not flow into the prediction targets). This shows that making the prediction targets stable and stale is the main cause of the improvement rather than the change in the objective due to the stop gradient. 

\begin{table}[t]
\centering \small

\begin{tabular}{l c c c c}
\toprule
Method & Predictor & Target parameters & $\beta$ & Top-1  \\
\midrule
\algo   & \checkmark & $\targetparams$           & 0 & $\bf{72.5}$ \\
        & \checkmark & $\targetparams$           & 1 & $70.9$ \\
        &            & $\targetparams$           & 1 & $70.7$ \\
        & \checkmark & $\mathrm{sg}(\netparams)$ & 1 & $70.2$ \\
\simclr &            & $\netparams$              & 1 & $69.4$ \\
        & \checkmark & $\mathrm{sg}(\netparams)$ & 1 & $70.1$ \\
        &            & $\mathrm{sg}(\netparams)$ & 1 & $69.2$ \\
        & \checkmark & $\netparams$              & 1 & $69.0$ \\
        & \checkmark & $\mathrm{sg}(\netparams)$ & 0 & $5.5$  \\
        & \checkmark & $\netparams$              & 0 & $0.3$  \\
        &            & $\targetparams$           & 0 & $0.2$  \\
        &            & $\mathrm{sg}(\netparams)$ & 0 & $0.1$  \\
        &            & $\netparams$              & 0 & $0.1$  \\
\bottomrule
\end{tabular}\vspace{0.25cm}
\captionsetup{width=.9\linewidth}
\caption{\label{tab:app_neg_term_1}Top-1 accuracy in \%, under linear evaluation protocol at 300 epochs, of  \mbox{intermediate} variants between \algo and \simclr (with caveats discussed in \Cref{app:simclr_baseline}). $\mathrm{sg}$ means stop gradient.}
\end{table}

\subsection{\simclr baseline of Section \ref{sec:contrast}}
\label{app:simclr_baseline}
The \simclr baseline in \Cref{sec:contrast} ($\beta = 1$, without predictor nor target network) is slightly different from the original one in~\cite{SimCLR}. 
First we multiply the original loss by $2\alpha$. For comparaison here is the original \simclr loss, 
\begin{align}
\text{InfoNCE}_{\netparams} \eqdef
 \frac{1}{B}\sum_{i=1}^B \frac{S_\netparams(v_i, v'_i)}{\alpha}\!-\! \frac{1}{B}\sum_{i=1}^B \ln\pa{\sum\limits_{j \neq i} \exp \frac{S_\netparams(v_i, v_j)}{\alpha} + \sum\limits_{j} \exp \frac{S_\netparams(v_i, v'_j)}{\alpha}}\cdot
\end{align}
Note that this multiplication by $2\alpha$ matters as the \LARS optimizer is not completely invariant with respect to the scale of the loss. Indeed, \LARS applies a preconditioning to gradient updates on all weights, except for biases and batch normalization parameters. Updates on preconditioned weights are invariant by multiplicative scaling of the loss. However, the bias and batch normalization parameter updates remain sensitive to multiplicative scaling of the loss. 

We also increase the original \simclr hidden and output size of the projector to respectively 4096 and 256. In our reproduction of \simclr, these three combined changes improves the top-1 accuracy at 300 epochs from $67.9\%$ (without the changes) to $69.2\%$ (with the changes).

\subsection{Ablation on the normalization in the loss function} 
\label{app:abla_norm}

\begin{figure}[ht!]
    \centering 
    \begin{subfigure}{0.65\linewidth}
        \centering
        \includegraphics[width=1.2\linewidth,bb=0 0 976 528]{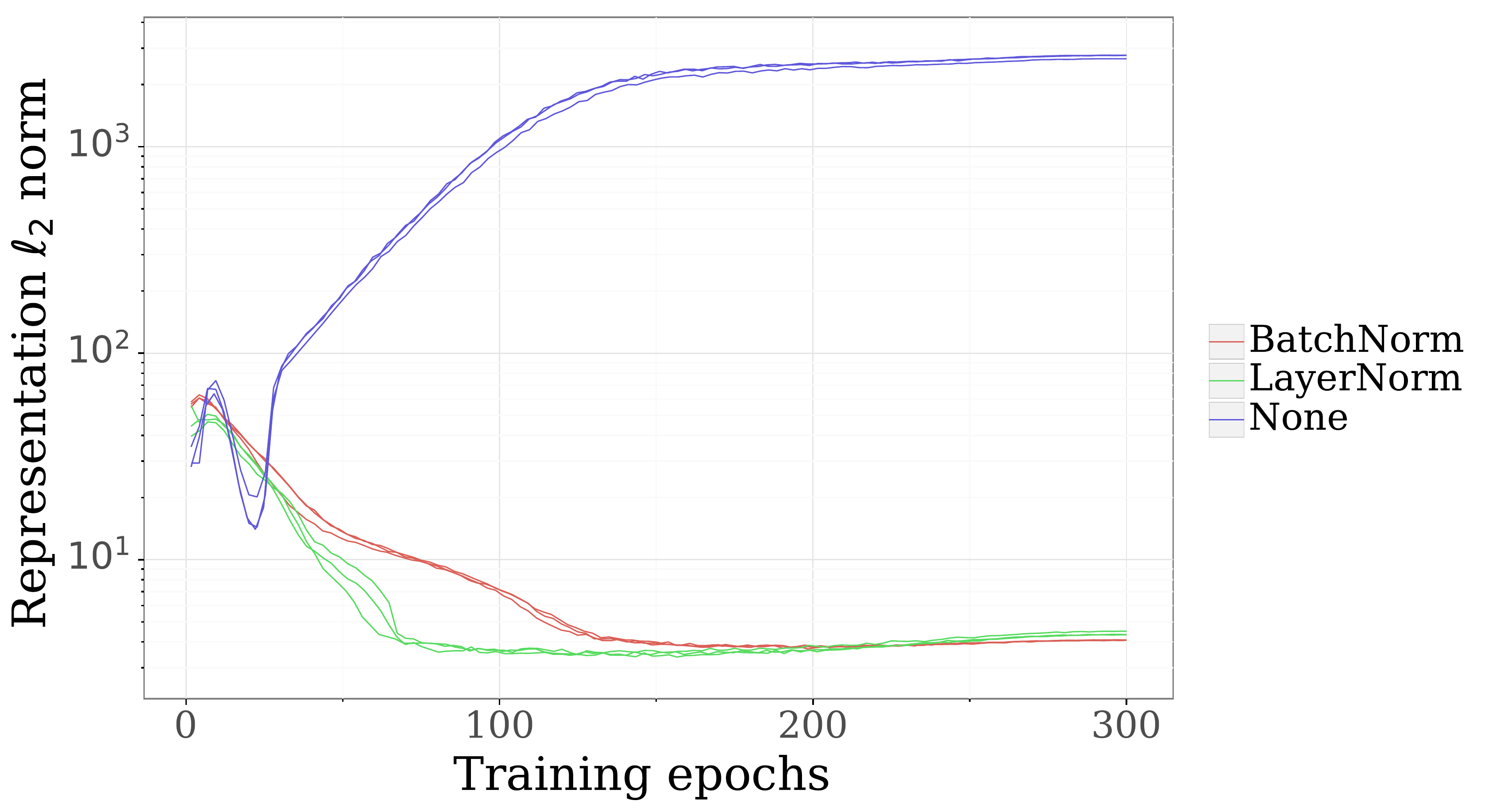}
        \caption{Representation $\ell_2$-norm}
        \label{fig:l2_norm_embedding_1}
    \end{subfigure}%

    \begin{subfigure}{0.65\linewidth}
        \centering
        \includegraphics[width=1.2\linewidth,bb=0 0 976 528]{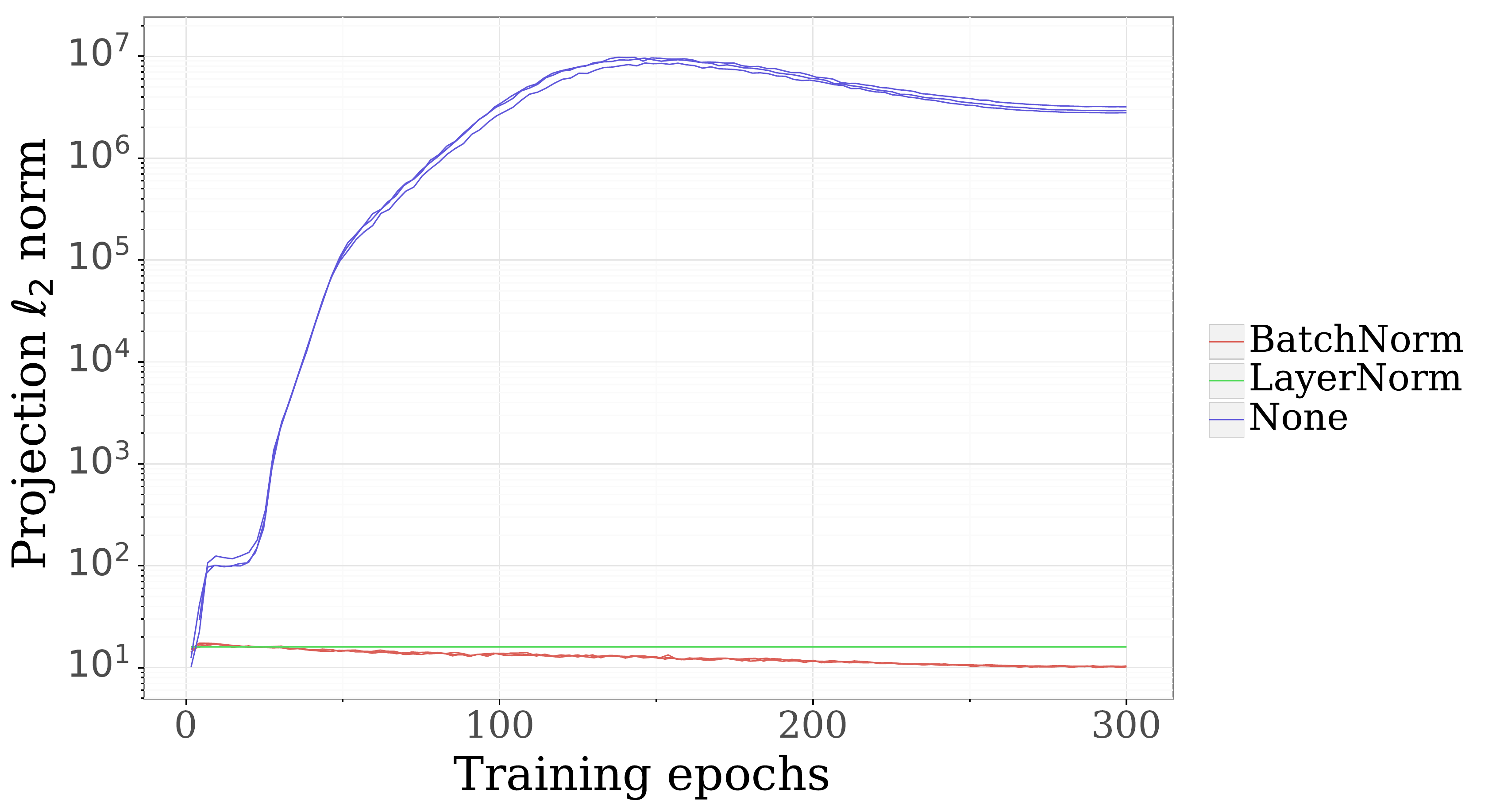}
        \caption{Projection $\ell_2$-norm}
        \label{fig:l2_norm_embedding_2}
    \end{subfigure}
    \caption{\label{fig:l2_norm_normalization} Effect of normalization on the $\ell_2$ norm of network outputs.}
\end{figure}

\vspace{0.1cm}
\begin{table}[ht!]
\centering 
\small
\begin{tabular}{l c c}
\toprule
Normalization & Top-1 & Top-5  \\
\midrule
$\ell_2$-norm & $\bf{72.5}$ & $\bf{90.8}$ \\
\LNt & $72.5 {\scriptstyle \pm 0.4}$ & $90.1$ \\
No normalization & $67.4$ & $87.1$ \\
\BNt & $65.3$ & $85.3$ \\
\bottomrule
\end{tabular}\vspace{0.25cm}
\captionsetup{width=.9\linewidth}
\caption{\label{tab:abla_norm}Top-1 accuracy in \% under linear evaluation \mbox{protocol} at 300 epochs for different normalizations in the loss.}
\end{table}

 \algo minimizes a squared error between the $\ell_2$-normalized prediction and target. We report results of \algo at 300 epochs using different normalization function and no normalization at all. More precisely,  given batch of prediction and targets in $\mathbb{R}^d$, $(p_{i}, t_{i})_{i\le B}$ with $B$ the batch size, \algo uses the loss function
  $\frac{1}{B}\sum_{i=1}^B \| n_{\ell_2}(p_i) - n_{\ell_2}(z_i)\|_2^2$ with 
  $n_{\ell_2} : x \rightarrow x/\|x\|_2$. 
  We run \algo with other normalization functions: 
  non-trainable batch-normalization and layer-normalization and no normalization. We divide the batch normalization and layer normalization by $\sqrt{d}$ to have a consistent scale with the $\ell_2$-normalization. We report results in \Cref{tab:abla_norm} where $\ell_2$, \LNt, no normalization and \BNt respectively denote using $n_{\ell_2}$, $n_\text{\BN}$, $n_\text{\LN}$ and $n_\text{\Id}$ with 

\begin{align*}
    {n_\text{\BN}^j}_{i}&: x \rightarrow \frac{x_{i}^j - \mu_\text{\BN}^j(x)}{\sigma_\text{BN}^j(x)\cdot\sqrt{d}}\CommaBin\quad 
    {n_\text{\LN}^j}_{i}: x \rightarrow \frac{x_{i}^j - {\mu_\text{\LN}}_i(x)}{{\sigma_\text{LN}}_i(x)\cdot\sqrt{d}}\CommaBin \quad 
    n_\text{\Id} : x \rightarrow x,\\
    \mu_\text{\BN}^j &: x\rightarrow \frac{1}{B}\sum_{i=1}^B x_i^j, \quad
    \sigma_\text{\BN}^j : x\rightarrow \sqrt{\frac{1}{B}\sum_{i=1}^B \left({x_i^j}\right)^2 - \mu_\text{\BN}^j(x)^2}, \quad\\
    {\mu_\text{\LN}}_i &: x\rightarrow \frac{1}{d}\sum_{j=1}^d x_i^j, \quad
    {\sigma_\text{\LN}}_i : x\rightarrow \frac{\|x_i - {\mu_{\LN}}_i(x)\|_2}{\sqrt{d}}
\end{align*}

When using no normalization at all, the projection $\ell_2$ norm rapidly increases during the first 100 epochs and stabilizes at around $3\cdot 10^6$ as shown in~\Cref{fig:l2_norm_normalization}. Despite this behaviour, using no normalization still performs reasonably well $(67.4\%)$. The $\ell_2$ normalization performs the best. 

\section{Training with smaller batch sizes}\label{app:smaller-batch}
The results described in \Cref{sec:experiments} were obtained using a batch size of $4096$ split over $512$ TPU cores. Due to its increased robustness, \algo can also be trained using smaller batch sizes without significantly decreasing performance. Using the same linear evaluation setup, \algo achieves $73.7\%$ top-1 accuracy when trained over $1000$ epochs with a batch size of $512$ split over $64$ TPU cores (approximately $4$ days of training). For this setup, we reuse the same setting as in \Cref{sec:method}, but use a base learning rate of $0.4$ (appropriately scaled by the batch size) and $\tau_\text{base} = 0.9995$ with the same weight decay coefficient of $1.5 \cdot 10^{-6}$.

\section{Details on Equation~\ref{eq:byol_minimize_cond_var} in Section~\ref{sec:intuition_byol}}
\label{app:thm-envelope}

In this section we clarify why \algo's update is related to \Cref{eq:byol_minimize_cond_var} from \Cref{sec:intuition_byol},
\begin{align}
    \nabla_\netparams\EE{\normtwo{q^\star(z_\netparams) - z'_\targetparams}^2} = \nabla_\netparams\EE{\normtwo{\EE{z'_\targetparams | z_\netparams} -z'_\targetparams }^2} = \nabla_\netparams \EE{ \sum_i \var(z'_{\targetparams, i} | z_\netparams)}. \tag{\ref{eq:byol_minimize_cond_var}}
\end{align}
Recall that $q^\star$ is defined as
\begin{align}
    q^\star \eqdef \argmin_q \EE{\normtwo{q(z_\netparams) - z'_\targetparams}^2}, \quad\text{where}\quad q^\star(z_\netparams) = \EE{z'_\targetparams | z_\netparams},
    \tag{\ref{eq:optimal_predictor}}
\end{align}
and implicitly depends on $\netparams$ and $\targetparams$; therefore, it should be denoted as $q^\star(\netparams, \targetparams)$ instead of just $q^\star$. For simplicity we  write $q^\star(\netparams, \targetparams)(z_\netparams)$ as $q^\star(\netparams, \targetparams, z_\netparams)$ the output of the optimal predictor for any parameters $\netparams$ and $\targetparams$ and input $z_\netparams$. 

\algo updates its online parameters following the gradient of \Cref{eq:byol_minimize_cond_var}, but considering only the gradients of $q$ with respect to its third argument $z$ when applying the chain rule. If we rewrite
\begin{align}
    \EE{\normtwo{q^\star(\netparams, \targetparams, z_\netparams) - z'_\targetparams}^2} = 
    \EE{L(q^\star(\netparams, \targetparams, z_\netparams), z'_\targetparams)},
\end{align}
the gradient of this quantity w.r.t.\,$\netparams$ is
\begin{align}
    \frac{\partial}{\partial \netparams}
    \EE{L(q^\star(\netparams, \targetparams, z_\netparams), z'_\targetparams)} = 
    \EE{\frac{\partial L}{\partial q}\cdot\frac{\partial q^\star}{\partial \netparams} + 
    \frac{\partial L}{\partial q}\cdot\frac{\partial q^\star}{\partial z}\cdot\frac{\partial z_\netparams}{\partial \netparams}},
\end{align}
where $\frac{\partial q^\star}{\partial \netparams}$ and $\frac{\partial q^\star}{\partial z}$ are the gradients of $q^\star$ with respect to its first and last argument. 
Using the envelope theorem, and thanks to the optimality condition of the predictor, the term $\EE{\frac{\partial L}{\partial q}\cdot\frac{\partial q^\star}{\partial \netparams}} = 0$. Therefore, the remaining term $\EE{\frac{\partial L}{\partial q}\cdot\frac{\partial q^\star}{\partial z}\cdot\frac{\partial z_\netparams}{\partial \netparams}}$ where gradients are only back-propagated through the predictor's input is exactly the direction followed by \algo. 

\section{Importance of a near-optimal predictor}
\label{app:importance_no_pred}
In this part we build upon the intuitions of \Cref{sec:intuition_byol} on the importance of keeping the predictor near-optimal. Specifically, we show that it is possible to remove the exponential moving average in \algo's target network (\emph{i.e.}, simply copy weights of the online network into the target) without causing the representation to collapse, provided the predictor remains sufficiently good.

\subsection{Predictor learning rate}
In this setup, we remove the exponential moving average (\emph{i.e.}, set $\tau = 0$ over the full training in \Cref{eq:tau}), and multiply the learning rate of the predictor by a constant $\lambda$ compared to the learning rate used for the rest of the network; all other hyperparameters are unchanged. As shown in \Cref{tab:pred_lr}, using sufficiently large values of $\lambda$ provides a reasonably good level of performance 
and the performance sharply decreases with $\lambda$
to 
$0.01\%$ top-1 accuracy (no better than random) for $\lambda = 0$.


To show that this effect is directly related to a change of behavior in the predictor, and not only to a change of learning rate in any subpart of the network, we perform a similar experiment by using a multiplier $\lambda$ on the \emph{predictor}'s learning rate, and a different multiplier $\mu$ for the \emph{projector}.
In \Cref{tab:pred_proj_lr}, we show  that the representation typically collapses or performs poorly when the predictor learning rate is lower or equal to that of the projector. As mentioned in Section~\ref{sec:intuition_byol}, we further hypothesize that one of the contributions of the target network is to maintain a near optimal predictor at all times.

\subsection{Optimal linear predictor in closed form}
Similarly, we can get rid of the slowly moving target network if we use a closed form optimal predictor on the batch, instead of a learned, non-optimal one. In this case we restrict ourselves to a linear predictor,
\begin{align*}
    q^\star = \argmin_{Q} \normtwo{Z_\netparams Q - Z'_\targetparams}^2= \pa{Z_\netparams\transpose Z_\netparams}^{-1} Z_\netparams\transpose Z'_\targetparams
\end{align*}
with $\normtwo{\cdot}$ being the Frobenius norm, $Z_\netparams$ and $Z'_\targetparams$ of shape $(B, F)$ respectively the online and target projections, where $B$ is the batch size and $F$ the number of features; and $q^\star$ of shape $(F, F)$, the optimal linear predictor for the batch. 

At $300$ epochs, when using the closed form optimal predictor, and directly hard copying the weights of the online network into the target, we obtain a top-$1$ accuracy of $fill$.


\begin{table}[ht]
\centering \small
\begin{tabular}{cr}
\toprule
   $\lambda$ & Top-$1$   \\
\midrule
$0$ & $0.01$  \\
                         $1$ & $5.5$  \\
                         $2$ & $62.8 {\scriptstyle \pm 1.5}$ \\
                        $10$ & $66.6$  \\
                        $20$ & $66.3 {\scriptstyle \pm 0.3}$ \\
\midrule
Baseline & $72.5$ \\
\bottomrule
\end{tabular}\vspace{0.25cm}
\captionsetup{width=.9\linewidth}
\caption{Top-$1$ accuracy at 300 epochs when removing the slowly moving target network, directly hard copying the weights of the online network into the target network, and applying a multiplier to the predictor learning rate.}
\label{tab:pred_lr}
\end{table}

\begin{table}[ht]
\centering \small
\begin{tabular}{ccrrrrr}
 \addlinespace[-\aboverulesep] 
 \cmidrule[\heavyrulewidth]{3-7}
        & & \multicolumn{5}{c}{$\lambda_{pred}$}  \\
        \cmidrule{3-7}
    & & $1$ & $1.5$ & $2$ & $5$ & $10$ \\
\cmidrule{2-7}
\multirow{5}{*}{$\mu_{proj}$}  & \multicolumn{1}{|c|}{$1$} & {\color{OliveGreen!4.783258594917787!BrickRed} $3.2 {\scriptstyle \pm 2.9}$} & {\color{OliveGreen!38.41554559043348!BrickRed} $25.7 {\scriptstyle \pm 6.6}$} & {\color{OliveGreen!90.88191330343795!BrickRed} $60.8 {\scriptstyle \pm 2.9}$} & {\color{OliveGreen!99.70104633781763!BrickRed} $66.7 {\scriptstyle \pm 0.4}$} & {\color{OliveGreen!100.0!BrickRed} $66.9$} \\
& \multicolumn{1}{|c|}{$1.5$} & {\color{OliveGreen!2.092675635276532!BrickRed} $1.4 {\scriptstyle \pm 2.0}$} & {\color{OliveGreen!13.751868460388637!BrickRed} $9.2 {\scriptstyle \pm 7.0}$} & {\color{OliveGreen!82.51121076233183!BrickRed} $55.2 {\scriptstyle \pm 5.8}$} & {\color{OliveGreen!91.92825112107623!BrickRed} $61.5 {\scriptstyle \pm 0.6}$} & {\color{OliveGreen!98.65470852017937!BrickRed} $66.0 {\scriptstyle \pm 0.3}$} \\
& \multicolumn{1}{|c|}{$2$} & {\color{OliveGreen!2.9895366218236172!BrickRed} $2.0 {\scriptstyle \pm 2.8}$} & {\color{OliveGreen!7.922272047832585!BrickRed} $5.3 {\scriptstyle \pm 1.9}$} & {\color{OliveGreen!23.617339312406575!BrickRed} $15.8 {\scriptstyle \pm 13.4}$} & {\color{OliveGreen!91.03139013452915!BrickRed} $60.9 {\scriptstyle \pm 0.8}$} & {\color{OliveGreen!99.1031390134529!BrickRed} $66.3$} \\
& \multicolumn{1}{|c|}{$5$} & {\color{OliveGreen!2.242152466367713!BrickRed} $1.5 {\scriptstyle \pm 0.9}$} & {\color{OliveGreen!3.736920777279521!BrickRed} $2.5 {\scriptstyle \pm 1.5}$} & {\color{OliveGreen!3.736920777279521!BrickRed} $2.5 {\scriptstyle \pm 1.4}$} & {\color{OliveGreen!30.642750373692074!BrickRed} $20.5 {\scriptstyle \pm 2.0}$} & {\color{OliveGreen!90.43348281016442!BrickRed} $60.5 {\scriptstyle \pm 0.6}$} \\
& \multicolumn{1}{|c|}{$10$} & {\color{OliveGreen!0.14947683109118085!BrickRed} $0.1$} & {\color{OliveGreen!3.139013452914798!BrickRed} $2.1 {\scriptstyle \pm 0.3}$} & {\color{OliveGreen!2.840059790732436!BrickRed} $1.9 {\scriptstyle \pm 0.8}$} & {\color{OliveGreen!4.185351270553064!BrickRed} $2.8 {\scriptstyle \pm 0.4}$} & {\color{OliveGreen!12.406576980568012!BrickRed} $8.3 {\scriptstyle \pm 6.8}$} \\
\bottomrule
\end{tabular}\vspace{0.25cm}
\captionsetup{width=.9\linewidth}
\caption{Top-$1$ accuracy at 300 epochs when removing the slowly moving target network, directly hard copying the weights of the online network in the target network, and applying a multiplier $\mu$ to the projector and $\lambda$ to the predictor learning rate. The predictor learning rate needs to be higher than the projector learning rate in order to successfully remove the target network. This further suggests that the learning dynamic of predictor is central to \algo's stability.}
\label{tab:pred_proj_lr}
\end{table}

\label{app:byol_sketch}
\begin{figure}[ht!]
    \centering
   \vskip -15cm
  \includegraphics[width=1.5\linewidth,trim={0cm 0cm 1cm 1cm},clip, bb=0 0 1140 1018]{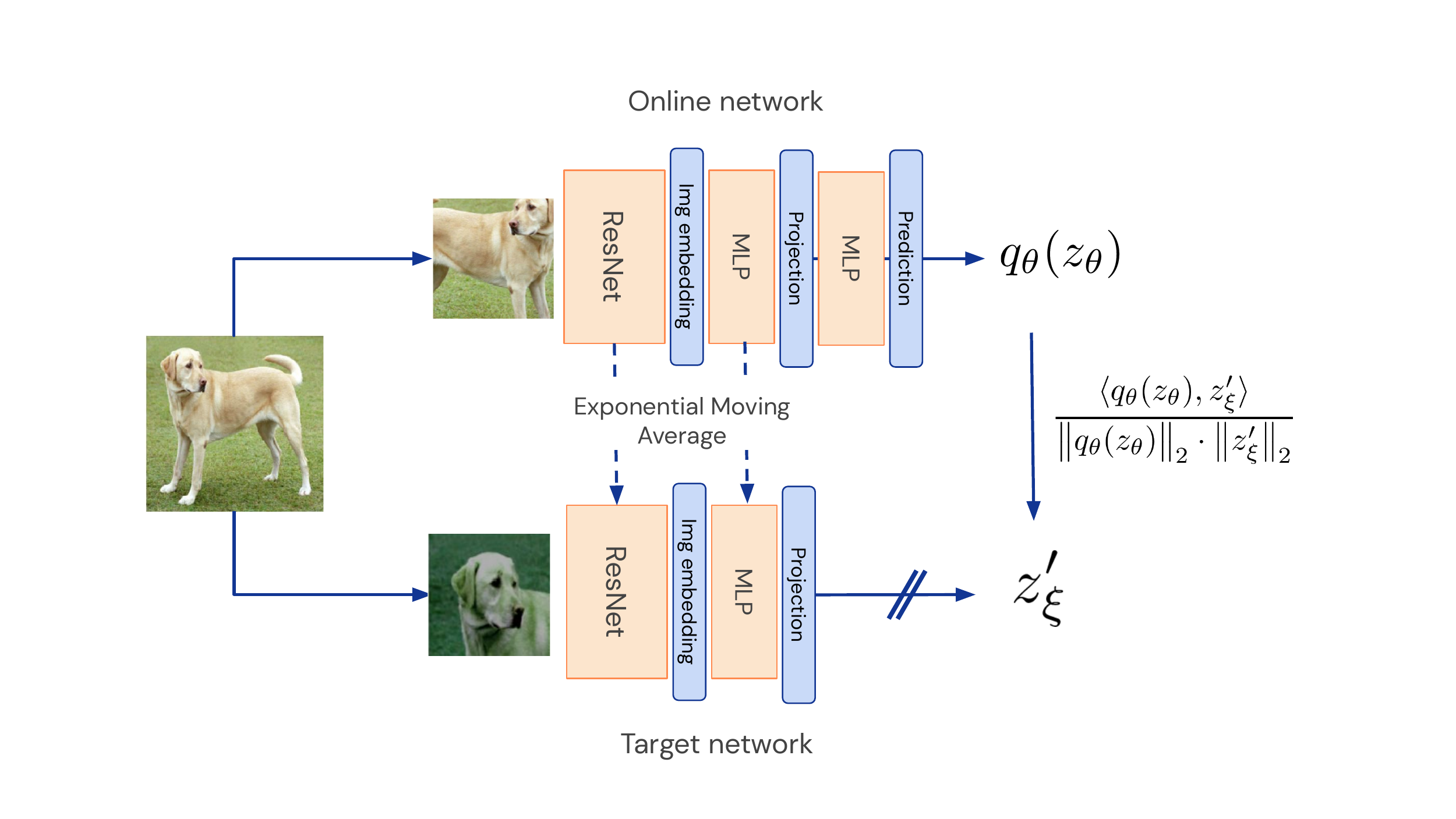}
  \caption{BYOL sketch summarizing the method by emphasizing the neural architecture.}
  \label{fig:BYOL_sketch}
\end{figure}

\clearpage


\section*{Supplementary material}

## J  BYOL pseudo-code in JAX

### J.1  Hyper-parameters

```python
HPS = dict(
    max_steps=int(1000. * 1281167 / 4096),  # 1000 epochs
    batch_size=4096,
    mlp_hidden_size=4096,
    projection_size=256,
    base_target_ema=4e-3,
    optimizer_config=dict(
        optimizer_name='lars',
        beta=0.9,
        trust_coef=1e-3,
        weight_decay=1.5e-6,
        # As in SimCLR and official implementation of LARS, we exclude bias
        # and batchnorm weight from the Lars adaptation and weightdecay.
        exclude_bias_from_adaption=True),
    learning_rate_schedule=dict(
        # The learning rate is linearly increase up to
        # its base value * batchisze / 256 after warmup_steps
        # global steps and then anneal with a cosine schedule.
        base_learning_rate=0.2,
        warmup_steps=int(10. * 1281167 / 4096),
        anneal_schedule='cosine'),
    batchnorm_kwargs=dict(
        decay_rate=0.9,
        eps=1e-5),
    seed=1337,
    )
```

### J.2  Network definition

```python
def network(inputs):
  """Build the encoder, projector and predictor."""
  embedding = ResNet(name='encoder', configuration='ResNetV1_50x1')(inputs)
  proj_out = MLP(name='projector')(embedding)
  pred_out = MLP(name='predictor')(proj_out)
  return dict(projection=proj_out, prediction=pred_out)


class MLP(hk.Module):
  """Multi Layer Perceptron, with normalization."""

  def __init(self, name):
    super().__init__(name=name)

  def __call__(self, inputs):
    out = hk.Linear(output_size=HPS['mlp_hidden_size'])(inputs)
    out = hk.BatchNorm(**HPS['batchnorm_kwargs'])(out)
    out = jax.nn.relu(out)
    out = hk.Linear(output_size=HPS['projection_size'])(out)
    return out

# For simplicity, we omit BatchNorm related states.
# In the actual code, we use hk.transform_with_state. The corresponding
# net_init function outputs both a params and a state variable,
# with state containing the moving averages computed by BatchNorm
net_init, net_apply = hk.transform(network)
```

## J.3 Loss function

```python
def loss_fn(online_params, target_params, image_1, image_2):
  """Compute BYOL's loss function.

  Args:
    online_params: parameters of the online network (the loss is later
      differentiated with respect to the online parameters).
    target_params: parameters of the target network.
    image_1: first transformation of the input image.
    image_2: second transformation of the input image.

  Returns:
    BYOL's loss function.
  """

  online_network_out_1 = net_apply(params=online_params, inputs=image_1)
  online_network_out_2 = net_apply(params=online_params, inputs=image_2)
  target_network_out_1 = net_apply(params=target_params, inputs=image_1)
  target_network_out_2 = net_apply(params=target_params, inputs=image_2)

  def regression_loss(x, y):
    norm_x, norm_y = jnp.linalg.norm(x, axis=-1), jnp.linalg.norm(y, axis=-1)
    return -2. * jnp.mean(jnp.sum(x * y, axis=-1) / (norm_x * norm_y))

  # The stop_gradient is not necessary as we explicitly take the gradient with
  # respect to online parameters only. We leave it to indicate that gradients
  # are not backpropagated through the target network.
  loss = regression_loss(online_network_out_1['prediction'],
                         jax.lax.stop_gradient(target_network_out_2['projection']))
  loss += regression_loss(online_network_out_2['prediction'],
                          jax.lax.stop_gradient(target_network_out_1['projection']))
  return loss
```

## J.4 Training loop

```python
def main(dataset):
  """Main training loop."""

  rng = jax.random.PRNGKey(HPS['seed'])
  rng, rng_init = jax.random.split(rng, num=2)
  dataset = dataset.batch(HPS['batch_size'])
  dummy_input = dataset.next()
  byol_state = init(rng_init, dummy_input)

  for global_step in range(HPS['max_steps']):
    inputs = dataset.next()

    rng, rng1, rng2 = jax.random.split(rng, num=3)
    image_1 = simclr_augmentations(inputs, rng1, image_number=1)
    image_2 = simclr_augmentations(inputs, rng2, image_number=2)
    byol_state = update_fn(
        **byol_state,
        global_step=global_step,
        image_1=image_1,
        image_2=image_2)

  return byol_state['online_params']
```

## J.5 Update function

```python
optimizer = Optimizer(**HPS['optimizer_config'])


def update_fn(online_params, target_params, opt_state, global_step, image_1,
              image_2):
  """Update online and target parameters.

  Args:
    online_params: parameters of the online network (the loss is  differentiated
      with respect to the online parameters only).
    target_params: parameters of the target network.
    opt_state: state of the optimizer.
    global_step: current training step.
    image_1: first transformation of the input image.
    image_2: second transformation of the input image.

  Returns:
    Dict containing updated online parameters, target parameters and
    optimization state.
  """
  # update online network
  grad_fn = jax.grad(loss_fn, argnums=0)
  grads = grad_fn(online_params, target_params, image_1, image_2)
  lr = learning_rate(global_step, **HPS['learning_rate_schedule'])
  updates, opt_state = optimizer(lr).apply(grads, opt_state, online_params)
  online_params = optix.apply_updates(online_params, updates)

  # update target network
  tau = target_ema(global_step, base_ema=HPS['base_target_ema'])
  target_params = jax.tree_multimap(lambda x, y: x + (1 - tau) * (y - x),
                                    target_params, online_params)
  return dict(
      online_params=online_params,
      target_params=target_params,
      opt_state=opt_state)


def init(rng, dummy_input):
  """BYOL's state initialization.

  Args:
    rng: random number generator used to initialize parameters.
    dummy_input: a dummy image, used to compute intermediate outputs shapes.

  Returns:
    Dict containing initial online parameters, target parameters and
    optimization state.
  """
  online_params = net_init(rng, dummy_input)
  target_params = net_init(rng, dummy_input)
  opt_state = optimizer(0).init(online_params)
  return dict(
      online_params=online_params,
      target_params=target_params,
      opt_state=opt_state)
```


\newpage

\end{document}